\pdfoutput=1

\documentclass[11pt]{article}

\usepackage[]{acl2023}

\usepackage{times}
\usepackage{latexsym}
\usepackage{graphicx}
\usepackage{adjustbox}
\usepackage{threeparttable}
\usepackage{CJKutf8}
\usepackage{multirow}
\usepackage{makecell}
\usepackage[T1]{fontenc}

\usepackage[utf8]{inputenc}
\usepackage{CJKutf8}
\usepackage{microtype}

\usepackage{float}
\usepackage{booktabs}
\usepackage[normalem]{ulem}
\useunder{\uline}{\ul}{}

\title{WYWEB: A NLP Evaluation Benchmark For Classical Chinese}

\author{Bo Zhou\textsuperscript{1,2}, Qianglong Chen\textsuperscript{1}, Tianyu Wang\textsuperscript{2}, Xiaomi Zhong\textsuperscript{2}, Yin Zhang\textsuperscript{1}\thanks{\ \ Corresponding Author: Yin Zhang.} \\
\textsuperscript{1}College of Computer Science and Technology, Zhejiang University, China \\
\textsuperscript{2}Xiaoniao.AI \\
        \texttt{\{zbo, chenqianglong, zhangyin98\}@zju.edu.cn} \\
        \texttt{\{zxm, wty\}@xiaoniao.ai} 
}

\begin{document}
\begin{CJK*}{UTF8}{gkai}
\maketitle
\begin{abstract}
To fully evaluate the overall performance of different NLP models in a given domain, many evaluation benchmarks are proposed, such as GLUE, SuperGLUE and CLUE. The field of natural language understanding has traditionally focused on benchmarks for various tasks in languages such as Chinese, English, and multilingual, however, there has been a lack of attention given to the area of classical Chinese, also known as "wen yan wen (文言文)", which has a rich history spanning thousands of years and holds significant cultural and academic value.\par
For the prosperity of the NLP community, in this paper, we introduce the WYWEB evaluation benchmark, which consists of nine NLP tasks in classical Chinese, implementing sentence classification, sequence labeling, reading comprehension, and machine translation. 
We evaluate the existing pre-trained language models, which are all struggling with this benchmark. 
We also introduce a number of supplementary datasets and additional tools to help facilitate further progress on classical Chinese NLU.
The github repository is https://github.com/baudzhou/WYWEB.
\end{abstract}

\section{Introduction}


Classical Chinese, as a written form of the Chinese language, had been widely used in the Confucian cultural circle, including China, Japan, Korea, Vietnam, etc ~\cite{Ye:2013, Nguyen:2020, Xu:95, Zhou:2009, Jin:2004}. As far as we know, there are about 400 million words, 3 million ancient articles have been passed down, covering literature, art, history, philosophy, etc, half of which are of great value ~\cite{Yin:2018}. 
However, in recent centuries, the use of classical Chinese has been gradually phased out and replaced by modern languages, resulting in increasing difficulty in comprehending it.
Therefore, it is necessary to introduce efficient NLP technology to process, understand, and research such literature.


While pre-trained language models such as BERT ~\cite{devlin-etal-2019-bert} and BERT-like models ~\cite{yang2019xlnet,dong2019unified,lan2019albert,liu2019roberta,he2020deberta,raffel2019exploring,wang2019structbert} have shown remarkable performance on English NLP benchmarks, including GLUE ~\cite{wang2018glue} and SuperGLUE ~\cite{NEURIPS2019_4496bf24}, there are also many efforts ~\cite{cui2020revisiting,wei2019nezha,cui2021preCBert} in Chinese NLP community, achieving significant improvement on modern Chinese NLP benchmark~\cite{xu2020clue, cui2018span, duan2019cjrc,cui-etal-2020-sentence}. 

However, since classical Chinese differs from modern Chinese in writing and grammar, these benchmarks can not be applied well to the studies in the classical Chinese domain.
In order to better adapt to the understanding of classical Chinese, many tasks and datasets are required to be redesigned, such as sequence labeling and sentence pair similarity. 
Meanwhile, due to the performance of the model being closely related to the pre-training corpus~\cite{qiu2020pre}, such as scale, language, domain, etc., the existing language models pre-trained on modern Chinese corpus can not adapt well to classical Chinese tasks.

Considering that previous studies ~\cite{ wangdongbo:sikubert2021, yang2021guwenunilm, Yasuoka:2022} for classical Chinese have typically evaluated models on few or different NLU tasks, it is difficult to compare the performances of these models. To facilitate such research in classical Chinese, it is necessary to design a standard classical Chinese NLP evaluation benchmark. 

In this paper, we introduce WYWEB (Wen Yan Wen Evaluation Benchmark), which will be open, and continually developed as much as we can. To evaluate the performance of the models of classical Chinese language representation, we create and refine nine tasks for different aspects of language understanding.

Specifically, considering the importance of breaks and pauses in sentences for the comprehension of classical Chinese, for sequence labeling, we design two novel tasks, including punctuation \emph{PUNC} and named entity recognition \emph{GLNER}, to evaluate word separation capability of pre-trained language models.
For sentence classification, we design three novel tasks, including text category classification \emph{GJC}, written time classification \emph{TLC}, and emotion classification of poems task \emph{FSPC}. 
For reading comprehension, we create a multiple choice task named \emph{WYWRC}. And on the other hand, the assessment of the natural language comprehension capability of a model through previous reading comprehension tasks is challenged by the extensive use of rare vocabulary and idiomatic expressions in classical Chinese.
Therefore, we design a novel reading comprehension task, \emph{IRC}, from the exam papers and idiom dictionary. Meanwhile, since machine translation of classical Chinese is also a problem of great concern, we design a novel \emph{WYWMT} task to study this topic. 
In addition, considering that some tokens in the classic Chinese language have the functions of prepositions, conjunctions and auxiliaries, and the same token has different meanings in different sentences, we design a new task, \emph{Xuci}, for token comparison. 
More details of these tasks are described in Section~\ref{sec:Tasks} and Appendix~\ref{app:data}. And we describe the principles we used to design tasks and the process of data collection in Section~\ref{sec:Overview}. 

Furthermore, to better understand the challenges provided by WYWEB, we build a baseline for each task and evaluate several pre-trained models released by the community. The experimental results demonstrate that current state-of-the-art methods are struggling with these tasks, which suggests that those tasks in WYWEB can constitute a useful test-bed for developing and comparing NLP systems for classical Chinese.

The contributions of our work are summarized as follows:
\begin{itemize}
    \item We propose and establish a novel benchmark for classical Chinese natural language understanding via redesigning, creating and collecting nine classical Chinese NLP tasks.
    \item To validate the challenge of this benchmark on existing pre-trained model models, we conduct a series of experiments with several baselines. Experimental results demonstrate these baselines are struggling with these novel tasks in classical Chinese.
    \item Finally, we build an online leaderboard and provide an evaluation tool set for further exploration, which will be publicly accessible as soon as possible.
\end{itemize}

\section{Related Work}
\subsection{Benchmarks for Pre-trained Language Model}
With the rise of the pre-training language model, pre-training a model on large corpus and fine-tuning them on downstream tasks becomes a general practice in the NLP community. 
To evaluate the ability of pre-trained language models in NLP tasks, several benchmarks are proposed for NLU tasks, such as SentEval \citep{conneau2018senteval}, GLUE \citep{wang2018glue} and SuperGLUE \citep{NEURIPS2019_4496bf24}, making existing models more comparable. 
For Chinese NLU, CLUE ~\citep{xu2020clue} benchmark is proposed with more than 10 tasks, including most NLP problems. 
To evaluate the ability of pre-trained language models in both natural language understanding and generation, CUGE ~\cite{yao2021cuge} is proposed, which is designed as a hierarchical framework via a multilevel scoring strategy. Meanwhile, to evaluate whether language models can learn a linguistic phenomenon of Chinese, ~\citet{xiang-etal-2021-climp} develops CLiMP which covers 9 major Mandarin linguistic phenomena. QuoteR ~\cite{qi-etal-2022-quoter} is designed for the evaluation of quote recommendation methods. Moreover, CBLUE ~\cite{zhang2021cblue} is a biomedical language understanding benchmark for Chinese, which mainly focuses on information extraction. 

However, compared to modern Chinese, there has been a lack of sufficient datasets and benchmarks for classical Chinese.
CCLUE~\footnote{https://cclue.top/} provides 5 NLU tasks for classical Chinese, including punctuation, NER, classification, sentiment recognition and retrieval between classical and modern Chinese. 
Unfortunately, this project was not finished and is no longer maintained.
Therefore, a new carefully designed benchmark for classical Chinese is very crucial for current studies.

\begin{table*}
  \centering

  \begin{tabular}{lllllll}
  \hline
  \textbf{Task} & \textbf{Train} & \textbf{Dev} & \textbf{Test} & \textbf{Description} & \textbf{Metric} & \textbf{Source}\\
  \hline
  PUNC           & 90k   & 20k & 20k  & Sequence labeling        & F1     & Authoritative Texts           \\
  TLC           & 28k   & 6k & 6k  & Sentence classification        & Accuracy     & Ancient prose      \\
  GJC            & 100k  & 20k & 20k  & Sentence classification  & Accuracy     & Daizhige \\
  XuCi           & 800   & 200 & 200  & Token similarity        & Accuracy   & Exam papers               \\
  WYWRC            & 3k    & 500  & 500   & Reading comprehension    & Accuracy    & Exam papers \\
  IRC            & 3k    & 1k  & 1k   & Reading comprehension    & Accuracy    & Exam papers \\
  WYWMT          & 20k   & 3k  & 3k   & Machine Translation      & BLEU   & online   \\
  \hline
    GLNER            & 80k   & 18k & 18k  & Sequence labeling        & F1     & \citet{GULIAN2020}   \\ 
      FSPC           & 3000  & 1000 & 1000  & Sentence classification & Accuracy   & THU-FSPC            \\
  \hline
  \end{tabular}
  \caption{The statistics of tasks in WYWEB, including the number of dataset, task description, evaluation metric and source. The datasets, except GLNER and FSPC, are created by us. }
  \label{tab:overview}
  \end{table*}

\subsection{Corpus Datasets for Classical Chinese}
The largest classical corpus dataset available is Daizhige (殆知阁)\footnote{https://github.com/garychowcmu/daizhigev20}, which contains about 3.3 billion tokens of classical Chinese literature, making classical Chinese corpus not low-resource. Most of pre-training related works use this dataset for model training.
Ancient Chinese Corpus (ACC)~\footnote{https://catalog.ldc.upenn.edu/docs/LDC2017T14/} dataset contains the word segmented, POS-tagged data of Zuozhuan (an ancient Chinese history classical book). This dataset is widely used in ancient Chinese studies. 
Recently, ~\citet{zinin-xu-2020-corpus} introduces an open source corpus of Twenty-Four Histories and some other ancient books. Meanwhile, FSPC ~\cite{10.1145/3459637.3481964} and CCMP ~\cite{li2021CCPM} are proposed for ancient poem understanding. While CUGE ~\cite{yao2021cuge} uses CCMP as a sub-task for classical poetry matching, in this work, we apply the FSPC dataset for poetry emotion recognition.

\subsection{Pre-trained Models for Classical Chinese}
\label{sec:ptmcc}

In Classical Chinese pre-trained language models, SikuBERT and SikuRoBERTa ~\citep{wangdongbo:sikubert2021} are pre-trained BERT/RoBERTa model on the Si Ku Quan Shu (Complete library in the Four Branches of Literature) corpus, and evaluated on 4 tasks, 
which are built from ACC dataset. Meanwhile, based on RoBERTa model, GuwenBERT~\footnote{https://github.com/ethan-yt/guwenbert} is pre-trained on Daizhige corpus with continuous training method and is evaluated on several NLU tasks. 
Other works ~\cite{HU2021SentSeg, Yu2021Seg, yang2021guwenunilm} also evaluate their models on different few NLP tasks.
However, since these models are not consistent in their evaluation tasks, to compare the natural language understanding ability of different pre-trained models in classical Chinese, a standard evaluation benchmark is required.

\section{WYWEB Overview}
\label{sec:Overview}

In this section, we introduce the principles and methods we applied during the construction process of WYWEB, and describe a brief overview of tasks in Table~\ref{tab:overview}. 
Firstly, we describe the process of task design and the principles we follow. Then, we introduce the data selection principles in Section~\ref{sec:corpusdata}. 
Finally, we provide a description of the leaderboard and evaluation toolkit. 


\subsection{Task Design Principles}
In this work, to assure that the benchmark can evaluate most aspects of pre-trained models and language phenomenons, we design evaluation tasks following best practices of other NLP benchmarks ~\citep{xu2020clue,yao2021cuge,NEURIPS2019_4496bf24,wang2018glue} and suggestions from experts.

Following the principles of~\citet{xu2020clue}, firstly, these tasks should vary in most aspects of NLP, including text classification, reading comprehension and machine translation, etc. 

Secondly, these tasks should be well-defined in the academic community and easily processed for corpus collection. 

Thirdly, they should be challenging but solvable. Finally, these tasks should be useful for follow-up studies and representative of classical Chinese natural language understanding tasks.

With the study of thousands of Chinese exam papers and requirements from academia and applications, we construct the evaluation tasks, covering most of the regular NLP tasks.

In addition to the regular tasks, we designed several tasks specifically for classical Chinese, i.e., punctuation of sentences without punctuation marks, comparison of confusing words and written period classification. 
These tasks will be introduced in Section~\ref{sec:Tasks} and Appendix~\ref{app:data}.

\subsection{Corpora Selection}
\label{sec:corpusdata}

\paragraph{Diversity Over Time}
Since classical Chinese has a very long history and evolves over time,  when designing tasks, we should choose texts that cover as many periods as possible. It is supposed that it is not reasonable enough to treat an isolated article as an independent task. 

\paragraph{Diversity Over Style} 
The stylistic theory is an important part of Chinese traditional literary theory. Since there are great differences between different styles, such as prose, parallel prose, poetry and so on, we believe that the benchmark should cover as many styles as possible.

For instance, \citet{wangdongbo:sikubert2021} evaluate their model on ACC corpus which is built on Zuo Zhuan. However, Zuo Zhuan was written by Zuo Qiuming in East Zhou Dynasty, so the text features of Zuo Zhuan are relatively simple and unable to test the models for a variety of linguistic phenomenons of classical Chinese. Therefore, in this work, we refine datasets like this and combine them into well-defined datasets to build uniform sample sets.

To evaluate an NLU ability for classical Chinese, it is natural to handle classical Chinese text as it is. However, dealing with raw classical Chinese text without sentence segmentation is extremely difficult, which leads to the task becoming unsolvable. Therefore, besides the PUNC task itself handling sentence segmentation of classical Chinese text, segmented texts are adopted for other tasks.

When selecting candidate corpora, we apply rules such as (1) having refined punctuation marks; (2) having more than 4 words in classification tasks; (3) being originally simplified Chinese character style preferred. 

\subsection{Data Collection and Ethical Concerns}
We collect data from as many channels as possible, i.e, open source projects, public websites, competition data and education institutions. Since classical Chinese sources are all works from many years ago, people could use the corpora for free. For other texts, if the copyright issue is concerned, we have been granted to use the data in this work. For example, GLNER is data from a competition for classical Chinese, we contact the owner by email and finally get licensed. 

Classical Chinese was officially and commonly used as a written language before recent times in East Asia, but now, modern languages have taken its place in these countries. In China, people learn classical Chinese at school but rarely use it in everyday life except for some poems and idioms. We could hardly collect any publicly available NLP datasets compared to modern Chinese. As a result, we design and create most of WYWEB datasets by ourselves. 

\subsection{Annotation}
\paragraph{Annotation Process} 
For different tasks, we apply different annotation processes. For PUNC, TLC and GJC tasks, the process is "data collecting, extracting, proofreading and final review". Specifically, since the data source is very important, these tasks are created using high quality and authoritative documents. Then paragraphs are sampled from every document with a certain small proportion, avoiding information leakage as much as possible. The annotators double-check every sample to ensure correctness of the computer work. Finally, domain experts review the whole work to get the result dataset.

For other tasks extracted from examination papers, the process is different because they need much more human workload. Given the papers (part of them are in PDF format or images), annotators copy or type the questions and do proofreading to assure the quality. Some new samples are also created by annotators to enrich the task. After the annotation work, the datasets are also sent to domain experts to carry out a final review.

\paragraph{Annotators} 
This is a community-based project, and most of the annotators and reviewers are volunteer students and scholars who are interested in classical Chinese or natural language processing. The authors take on the remaining annotation tasks.

It takes us a long time to create this benchmark, so the annotators vary during the process. When selecting annotators, we make sure they have got a good language score in national college entrance examination and are familiar with classical Chinese. Some rules annotators following are: (1) dropping out confusing sentences; (2) double-checking rarely used words and dropping out sentences with uncertain rarely used words; (3) removing unnecessary symbols except specified punctuation marks. To ensure the quality of the datasets, we asked some domain experts to do the final review. 

For instance, Buddhist scriptures in classical Chinese are kind of important documents. As Buddhist scriptures are generally collated and proofread by Buddhist believers, the correctness of these texts is relatively reliable. We apply many Buddhist scripture texts in our tasks, hence the need for a review of scholars from the Buddhist Academy.

\paragraph{Quality checks for WYWMT} Translating classical Chinese sentences to modern Chinese is challenging. We follow \citet{guzman2019flores} to filter texts collected from the internet. In addition, because classical Chinese sentences are usually short, the limitation of sample length is set to 5 to 200 characters.

\subsection{Toolkit and Leaderboard}

For the evaluation toolkit, we provide scripts implemented using PyTorch ~\cite{paszke2019pytorch} and transformers ~\cite{wolf2020transformers}, which can help the followers evaluate their models easily. Otherwise, they can upload their models to Hugging Face Model Hub~\footnote{https://huggingface.co/} and contact us for the evaluated results. This toolkit is also released on WYWEB repository. Furthermore, we provide a leaderboard for the community to present the performance of each model. The leaderboard includes a general list and a sub list of each task. 
The results will be updated soon after the submission of models. As shown in Section~\ref{leaderboard}, we provide some details and screen-shots of the leaderboard.

\section{Tasks}
\label{sec:Tasks}
In this section, we describe tasks and datasets designed for specific aspects of classical Chinese NLP. These datasets, except GLNER and FSPC, are firstly created by ourselves. 
More details are shown in Appendix~\ref{app:data}. 

\subsection{Sequence Classification Tasks}
\paragraph{GJC}
This task aims to work on the problem of ancient book classification which has been discussed since ancient times.
We select a proportion of text from each category of the Daizhige project and divide them into the training set and evaluation set, where each sample is a selected paragraph from an article or book, ranging from a few dozen to hundreds of characters in length.
More details are shown in Appendix~\ref{det:gjc}.






\paragraph{TLC} 
This task is to identify the written time of ancient books.
We create the TLC dataset where each sample has a coarse-grained label (period) and a fine-grained label (Dynasty) forming a hierarchical structure. Similar to GJC task, each sample is a paragraph selected from ancient literature. More details are shown in Appendix~\ref{det:tlc}.

\paragraph{FSPC}
FSPC (Fine-grained Sentiment Poetry Corpus) is an emotion recognition task for ancient rhythmic poetry, created by THUAIPoet (九歌) group ~\cite{10.1145/3459637.3481964}. Sentiments are annotated into 5 classes, i.e. negative, implicit negative, neutral, implicit positive, and positive. THUAIPoet designs a reasonable annotation mechanism to ensure annotations follow similar standards during the work process. See Appendix~\ref{det:fspc} for details.









\subsection{Sequence Labeling Tasks}
\paragraph{PUNC} 
This task is designed to perform text segmentation, i.e., adding punctuation marks to continuous ancient Chinese texts. The dataset uses some fairly authoritative texts, including historical records and Buddhist scriptures, to construct the samples. In order to reduce the complexity of the research, only a few commonly used punctuation marks are used. Each sample consists of the original text and its corresponding label sequence. Furthermore, the samples are paragraphs containing several sentences. See Appendix~\ref{det:punc} for details.








\paragraph{GLNER}
GLNER is a named entity recognition task created by ~\citet{GULIAN2020}. Texts of the dataset are selected from ancient books and some other relevant literature. There are two kinds of entities in this dataset, i.e., classical book name and other which includes human name, location name, etc. Since the entity category is of coarse grain size, it is expected to implement new labeling work to refine this dataset in the future. See Appendix~\ref{det:glner} for details.








\subsection{Sentence Pair Tasks}

\paragraph{XuCi}
This task is designed to determine whether two function words in a sentence pair have the same meaning and usage. The words to be compared in the samples generally consist of one or two single characters. Each sample includes fields such as the pair of sentences, the position of the function words in the sentence, and a label indicating whether they are the same or not (True or False). See Appendix~\ref{det:xuci} for details.




    

\subsection{Reading Comprehension Tasks}
\paragraph{WYWRC}
Similar to the RACE dataset \cite{lai2017race,sun2019probing}, this task involves providing a classical Chinese paragraph, a question, and selecting the best answer from among four options. This task is quite challenging, as the model must possess a proficient understanding of both modern and classical Chinese. For analysis purpose, we separate the samples into 10 types according to their questions and answers. More details are shown in Appendix~\ref{det:wywrc}.









\paragraph{IRC} 
Considering idiom comprehension is a very important part of classical Chinese learning, to evaluate the idiom comprehension ability of the model, we design and collect the IRC dataset. In IRC, given an idiom and its origin (most are in classical Chinese), the model is required to select the best explanation from four options. 
See Appendix~\ref{det:irc} for details.







\subsection{Sequence to Sequence Tasks}

\paragraph{WYWMT} 
Machine translation of classical Chinese is a problem of great concern. 
This task is used to evaluate whether pre-trained models can effectively improve the performance of machine translation models for classical Chinese. 
Due to the limited number of samples, we only use WYWMT dataset for evaluation rather than fine-tuning.
Furthermore, we separate this task from others and make a stand-alone leaderboard. See Appendix~\ref{det:wywmt} for details.


\section{Baselines}
\label{sec:Baselines}

\subsection{Baseline Implementation}

\begin{figure}[!t]
\centering
  \includegraphics[scale=0.4]{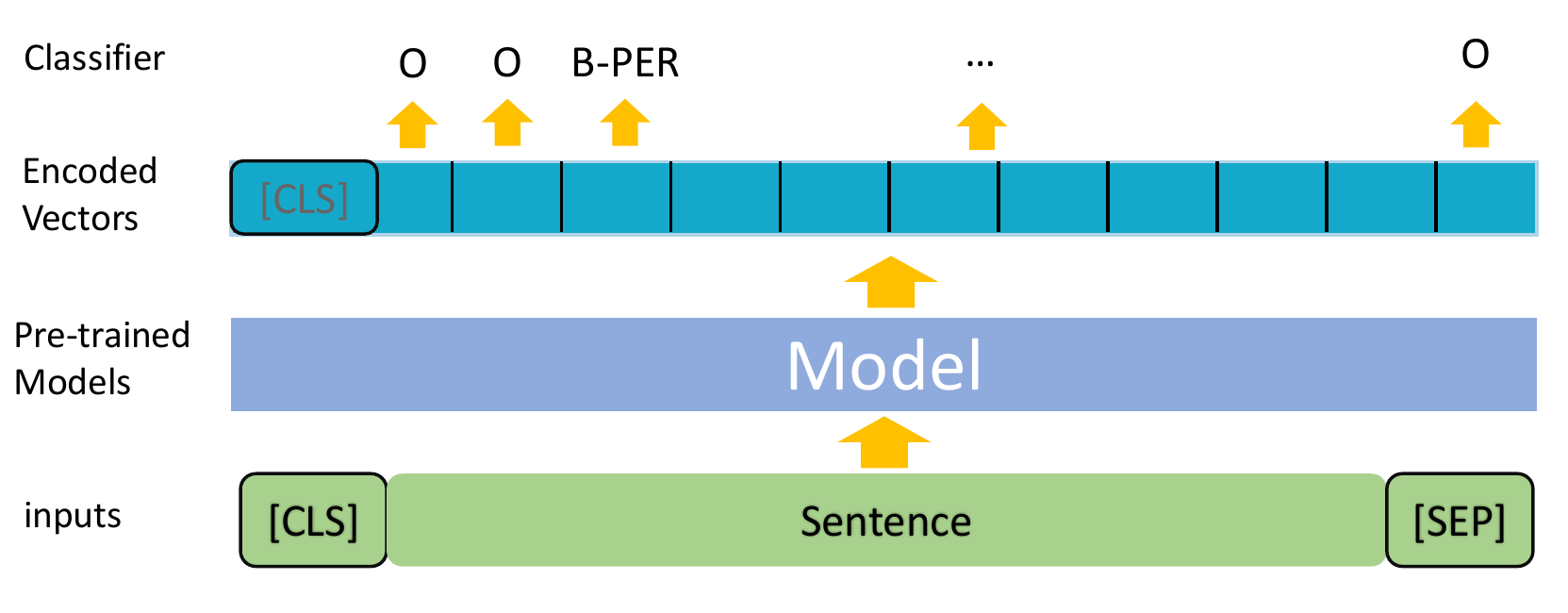}
  \caption{Implementation of Sequence Labeling tasks.}
  \label{basline:SL}
\end{figure}

\paragraph{Sequence Labeling} We get hidden states from the last layer of the model encoder, and pass them to a classifier to get sequence labels. See Figure~\ref{basline:SL}.

\begin{figure}[!t]
\centering
  \includegraphics[scale=0.4]{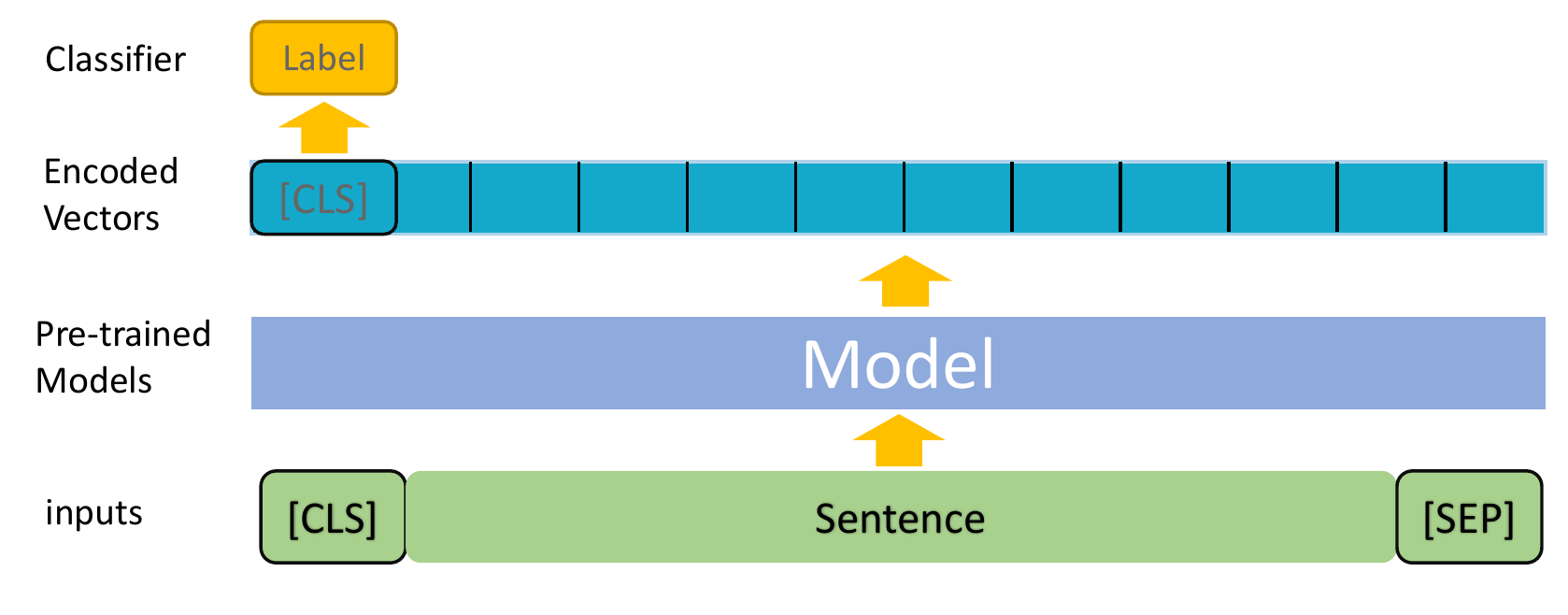}
  \caption{Implementation of Sentence Classification tasks.}
  \label{basline:SC}
\end{figure}

\paragraph{Sentence Classification} We get pooled output of model encoder, i.e., hidden state of [CLS] token, and pass it to a classifier to get sequence labels.
See Figure~\ref{basline:SC}.

\begin{figure}[!t]
\centering
  \includegraphics[scale=0.4]{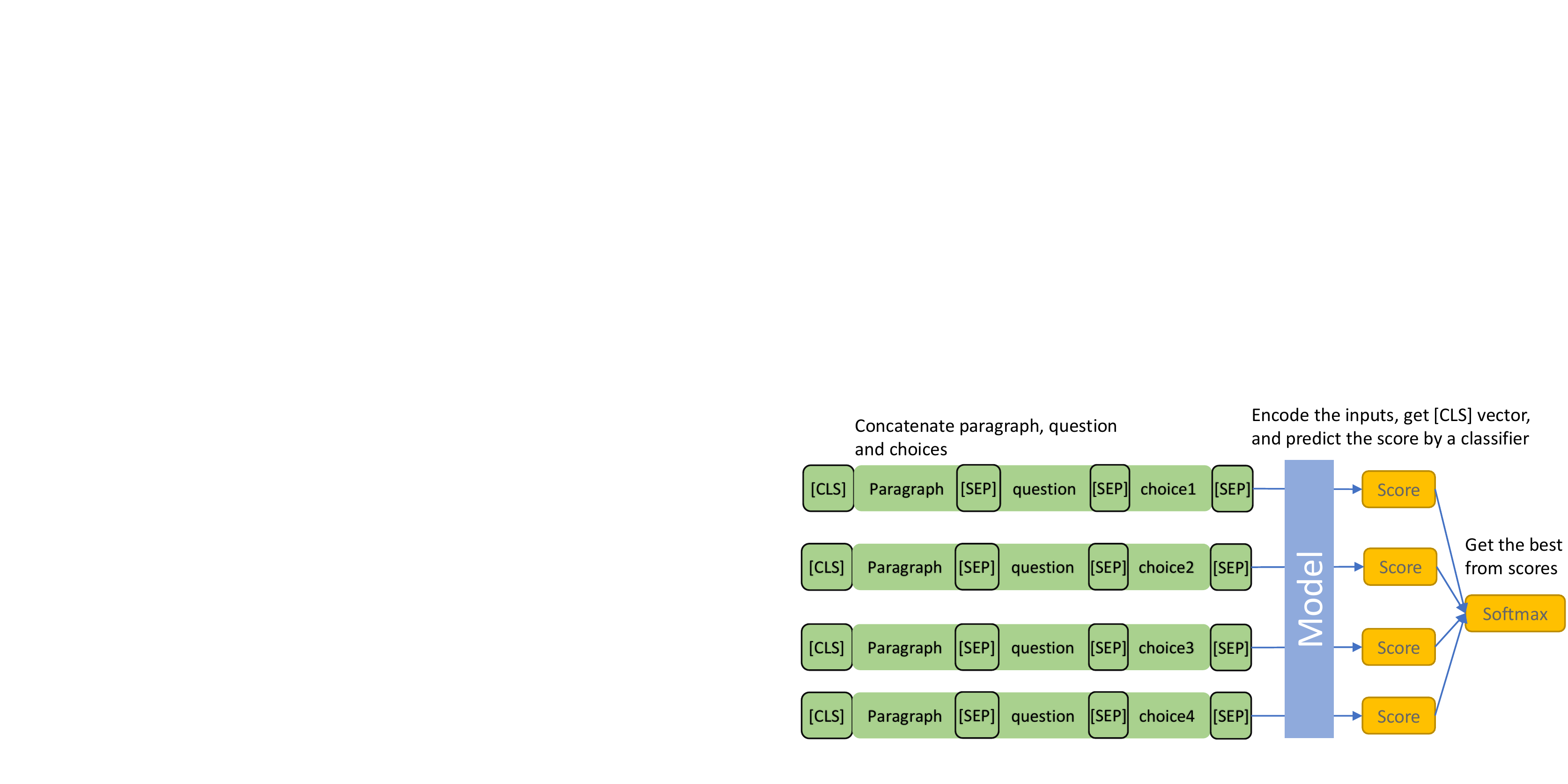}
  \caption{Implementation of Reading Comprehension tasks.}
  \label{basline:RC}
\end{figure}

\paragraph{Reading Comprehension} We encode each option concatenated with paragraph-question, pass the hidden states to a shared classifier to get prediction score and choose the best as final answer.
See Figure~\ref{basline:RC}.

\begin{figure}[!t]
\centering
  \includegraphics[scale=0.4]{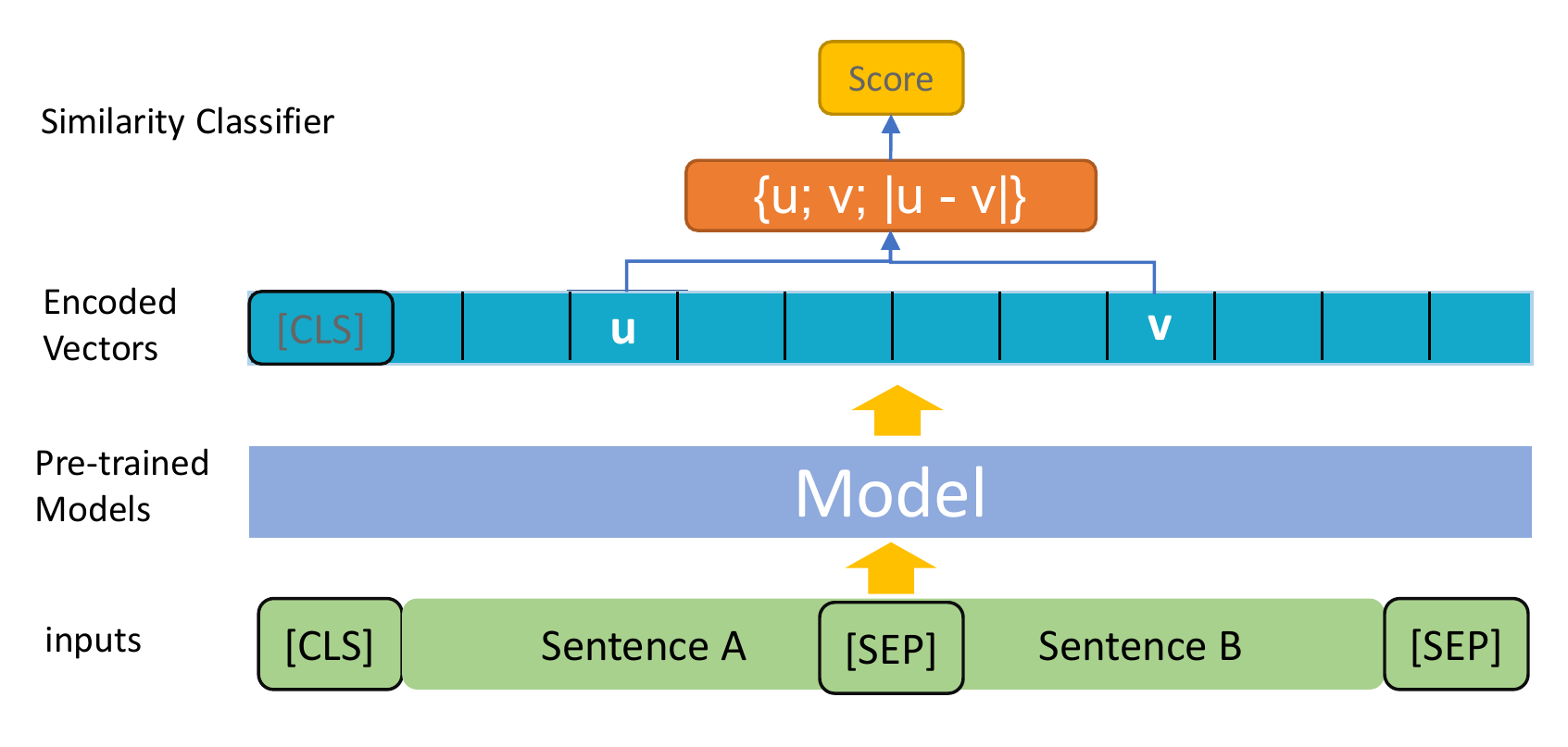}
  \caption{Implementation of Token Similarity tasks.}
  \label{basline:TC}
\end{figure}

\paragraph{Token Similarity} Similar to sentence classification, we encode sentence pairs and get the hidden state of the corresponding token, then we use \{$\mathbf{u}$ ; $\mathbf{v}$; |$\mathbf{u}$  - $\mathbf{v}$|\} to represent the similarity score, where we mark the vectors as $\mathbf{u}$ and $\mathbf{v}$.
See Figure~\ref{basline:TC}.

\paragraph{Machine Translation}
We implement this task as sentence pair with a prefix attention mask to adapt BERT-like models. To save inference time cost, the sequence output of the target sentence is greedy decoded. Note that this implementation is untypical for the sequence-to-sequence models, and is just used to reflect the capabilities of the model itself.

All the experiments are implemented using PyTorch \cite{paszke2019pytorch}. 
\begin{table*}[]
\centering
\begin{adjustbox}{width=1\textwidth}
\begin{tabular}{lccclccclclcc}
\hline
\multirow{2}{*}{\textbf{Models}} &
  \multirow{2}{*}{\textbf{Avg.}} &
  \multicolumn{2}{c}{\textbf{Sequence Labeling}} &
   &
  \multicolumn{3}{c}{\textbf{Sentence Classification}} &
   &
  \textbf{Token Sim.} &
   &
  \multicolumn{2}{c}{\textbf{Reading Comp.}} \\ \cline{3-4} \cline{6-8} \cline{10-10} \cline{12-13} 
                  &               & \textbf{PUNC}          & \textbf{GLNER}         &  & \textbf{GJC}           & \textbf{FSPC}          & \textbf{TLC}           &  & \textbf{Xuci}         &  &  \textbf{WYWRC}  & \textbf{IRC}           \\ \hline
Human             & 88.0          & 92.4          & 94.3          &  & 90.3          & 80.0          & 89.0          &  & 85.3          & & 80.0 & 92.3          \\ \hline
DeBERTa-base      & \textbf{75.9} & \textbf{83.3} & \textbf{86.7} &  & \textbf{85.2} & 61.1          & 86.7          &  & 72.4         &  & \textbf{45.1} & 86.7          \\ \hline
GuwenBERT-base    & 72.9          & 82.5          & 82.8          &  & 84.8          & 61.3          & 85.1          &  & 71.7          & & 28.0 & 86.8          \\
GuwenBERT-large   & 75.6          & 83.1          & 86.1          &  & 84.9          & 58.5          & \textbf{87.6} &  & 73.4 & & 44.4 & \textbf{87.8} \\
GuwenBERT-base-fs & 74.6          & 82.9          & 84.8          &  & 84.2          & 61.0          & 86.7          &  & 70.0         &  & 42.1 & 85.3          \\ \hline
RoBERTa-CCBC      & 74.5          & 82.5          & 84.7          &  & 84.5          & 59.5          & 85.0          &  & 73.2         &  & 40.7 & 86.1          \\
RoBERTa-CCLC      & 75.3          & 82.8          & 86.1          &  & 84.7          & 58.6          & 87.1          &  & \textbf{74.9}  & & 41.0  & 86.9          \\ \hline
SikuBERT          & 73.7          & 80.8          & 82.8          &  & 82.2          & 60.9          & 82.4          &  & 70.4         &  & 44.0 & 85.8          \\
SikuRoBERTa       & 73.5          & 81.4          & 82.8          &  & 82.5          & \textbf{62.2} & 83.8          &  & 68.5 &  & 41.0 & 85.8          \\ \hline

RoBERTa-wwm-ext   & 72.1          & 78.8          & 79.8          &  & 81.3          & 59.2          & 78.3          &  & 71.0         &  & 42.1 & 86.2          \\ \hline
\end{tabular}
  \end{adjustbox}

\caption{The results of baselines on the NLU tasks of WYWEB benchmark.}
\label{tab:results}
\end{table*}

\subsection{Pre-trained Models}

\paragraph{GuwenBERT} GuwenBERT has three versions, including GuwenBERT-base, GuwenBERT-large, GuwenBERT-fs-base. While GuwenBERT-base and GuwenBERT-large are trained based on RoBERTa-wwm-ext ~\cite{cui-etal-2021-pretrain}, a modern Chinese pre-trained model, and then continue trained on classical Chinese corpus, GuwenBERT-fs-base is trained purely on classical Chinese corpus.


\paragraph{RoBERTa-classical-chinese} RoBERTa-classical-chinese has two versions, including RoBERTa-classical-chinese-base-char (RoBERTa-CCBC), RoBERTa-classical-chinese-large-char (RoBERTa-CCLC).
This is a RoBERTa model pre-trained on classical Chinese texts, derived from GuwenBERT-base.
Character-embeddings are enhanced into traditional/simplified characters \cite{Yasuoka:2022}. 

\paragraph{SikuBERT, SikuRoBERTa}
These models are pre-trained on the verified high-quality “Siku Quanshu” ~\cite{wangdongbo:sikubert2021}. Note that these two models are pre-trained on traditional Chinese. In the fine-tuning stage, we convert simplified Chinese corpus into traditional Chinese.

\paragraph{DeBERTa-base} Based on the structure of DeBERTa ~\cite{he2020deberta}, we pre-trained the model on DaiZhiGe corpus from scratch.

\paragraph{RoBERTa-wwm-ext}
This model is trained with BERT (RoBERTa) structure ~\cite{cui-etal-2021-pretrain} and whole word masking.

Note that there are not as many pre-training models of classical Chinese as modern Chinese.
We collect all models of classical Chinese accessible to evaluate and take them as baselines. 
More details of these models can be found in Appendix~\ref{app:det_models}.

\subsection{Experiment Setting}

For the evaluation, we fine-tune the pre-trained models mentioned above.
For each task, we train 3 runs, and the model with the best development score is used for testing. 
When the learning rate decreases to a specified small value or the performance does not improve for 5 evaluations, the training is stopped. More details of hyper-parameters are shown in Appendix~\ref{app:hp}.

\subsection{Human Performance}
For all tasks, we evaluate human performance following the principle of SuperGLUE: extract 30 samples in the training phase, and then sample 100 items from the test set in the testing phase.
We collect test results from three annotators and calculate the human performance.
The annotators are all college students majoring in ancient Chinese.
The results of human performance are shown in Table~\ref{tab:results} and Table~\ref{tab:resultsMT}. More details are shown in Appendix~\ref{app:human}.
\begin{table*}
    \centering

    \begin{threeparttable}
      \begin{tabular}{@{}lccccccc@{}}
        \toprule
        \textbf{Model} & \textbf{BLEU} & \textbf{chrF2} & \textbf{TER\tnote{*}} & \textbf{ROUGE-1} & \textbf{ROUGE-2}      & \textbf{ROUGE-L}  \\ \midrule
        Human & 45.6 & 44.2 & 34.4 & 77.4 & 50.7 & 76.2 \\ \hline
        guwenbert-base & \textbf{40.1} & \textbf{38.1} & 37.5 & \textbf{72.5} & \textbf{46.0} & \textbf{70.3} \\
        guwenbert-large & 38.8 & 37.2 & 38.1 & 70.1 & 43.7 & 67.7\\
        guwenbert-base-fs & 36.3 & 35.2 & 39.2 & 68.3  & 41.2   & 65.7 \\\hline
        roberta-CCBC & 39.1 & 37.1  & 36.8 & 71.4 & 44.9 &69.3\\
        roberta-CCLC & 39.8 & 38.0 & 36.4 & 71.6 & 45.3 & 69.3 \\ \hline
        SikuBERT & 38.8 & 36.2  & 37.9 & 72.0 & 45.5 & 69.8 \\
        SikuRoBERTa & 39.1 & 36.5 & 37.7 & 72.2 & 45.7 & 70.0 \\\hline
        DeBERTa-base & 39.5 & 37.8 & \textbf{35.9} & 71.9 & 44.2 & 68.7 \\
        Roberta-wwm-ext & 38.0 & 35.8 & 39.1 & 69.9 & 43.2 & 66.7 \\ \bottomrule
        \end{tabular}

        \begin{tablenotes}
            \footnotesize
            \item[*] Translation Edit Rate
        \end{tablenotes}
        
        \end{threeparttable}

\caption{The results of baselines on WYWMT task of WYWEB benchmark. }
\label{tab:resultsMT}
\end{table*}

\subsection{Benchmark Results \& Analysis}
\label{sec:results}
As shown in Table~\ref{tab:results}, we present the performance of existing baseline models in classical Chinese NLU tasks. Since evaluation metrics of sequence-to-sequence tasks are different from NLU tasks, as shown in Table~\ref{tab:resultsMT}, we evaluate each model on WYWMT task independently with several metrics, including BLEU, chrF2, TER and ROUGE. 

From the results, it can be seen that some regular patterns, i.e. "the bigger (model scale and batch size), the better"; "the more (data and train steps), the better" appear as described in other experiments.

DeBERTa-base \cite{he2020deberta} performs best on this benchmark showing that the model structure and training strategy are both effective. 
Note that this model is pre-trained just according to the default settings of DeBERTa V2 English version without the convolution layer and purely on classical Chinese.
Meanwhile, some techniques that have obvious effects in Chinese are not used, such as Whole Word Masking \cite{cui-etal-2021-pretrain}, etc.
All models pre-trained on classical Chinese get better scores than chinese-roberta-wwm-ext \cite{cui-etal-2021-pretrain} pre-trained on modern Chinese corpus. Similarly, models trained on both classical Chinese and modern Chinese perform better on tasks involving both scripts, such as WYWRC, IRC and GLNER.


For FSPC task, composed of ancient Chinese rhythmic poems, SikuRoBERTa ~\cite{wangdongbo:sikubert2021} performs the best, which is pre-trained with a high-quality classical Chinese corpus of Si Ku Quan Shu rather than on Daizhige. 

Since poems are different from general texts, the models could learn a better representation of ancient words on Si Ku Quan Shu instead of on Daizhige.
The two large models yield similar scores to DeBERTa-base but much better than other smaller ones, however, the parameter size of the large model is 3 times larger than that of DeBERTa-base.

On WYWMT task, GuwenBERT-base achieves the best score with its pre-training strategy, which initializes the parameters of the transformer model from a pre-trained model and trains the model via freezing encoder layers to translate modern Chinese knowledge to classical and updates all parameters of the model. With the strategy, the model could learn a good representation of both modern and classical Chinese and achieve the best performance on translation tasks.

Compared with human performance, all the models have a big gap with the artificial results, especially on tasks WYWRC, XuCi, and IRC, which require a lot of implicit knowledge.

One limitation of our evaluation is, the models we collected are all BERT or RoBERTa style and are lacking some variety. Furthermore, the models we evaluated maybe not achieve the best score in this baseline due to differences among them. However, they are fine-tuned with similar hyper-parameters, so that the results are comparable as expected.

\subsection{Task Probing}
According to scores in Table~\ref{tab:results}, we choose the most challenging task, WYWRC for further exploration and analysis. 
As shown in details of Table~\ref{tab:wywrcType} in Appendix~\ref{det:wywrc}, this task has a variety of questions, which is different from traditional machine reading comprehension tasks, and is more difficult for existing models. 

\begin{figure}[!t]
\centering
  \includegraphics[scale=0.4]{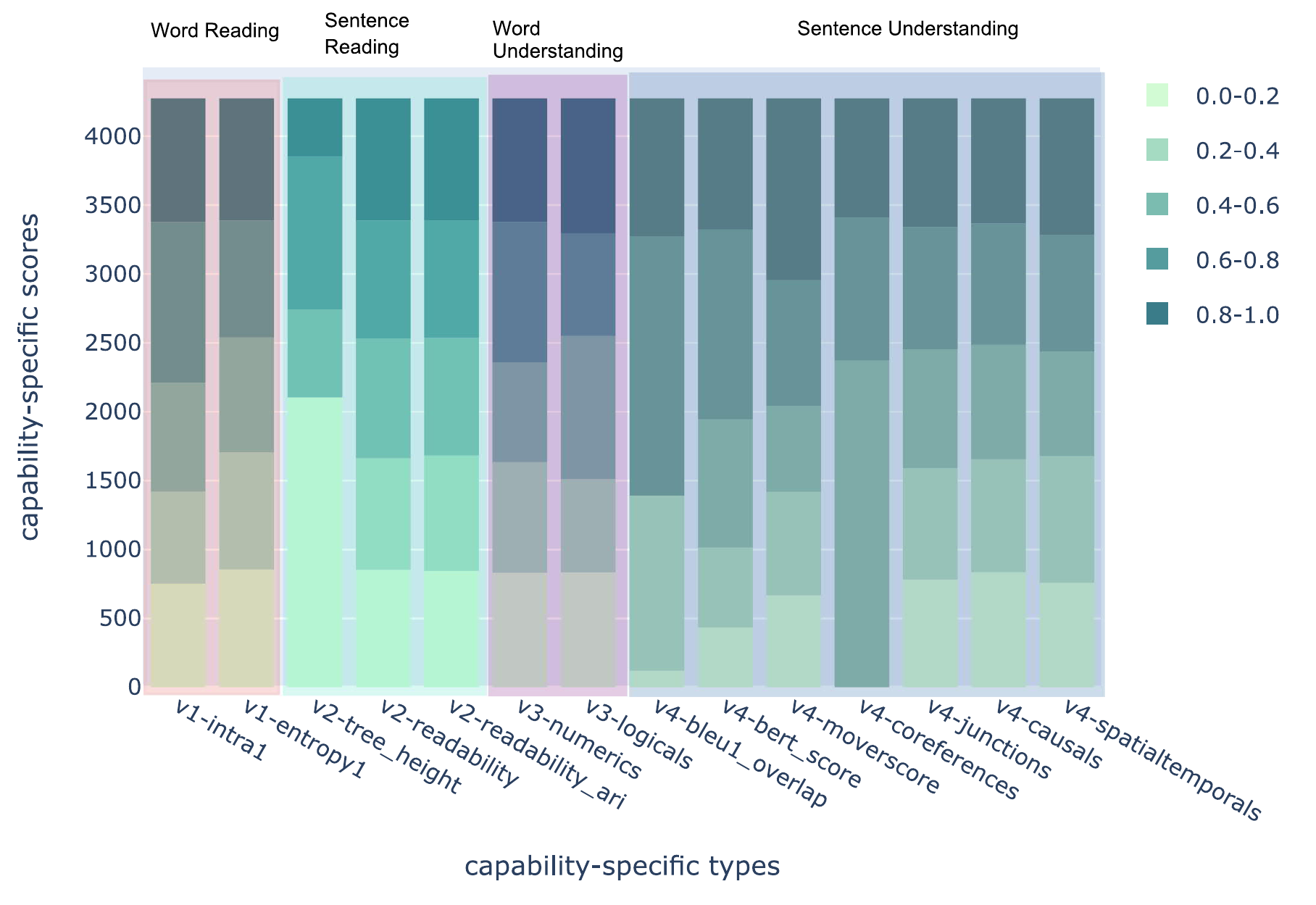}
  \caption{Competency assessment of MRC capabilities. The numbers in the legend, for example "0.2-0.4", indicate capability-speciﬁc value score.}
  \label{wywrc:css}
\end{figure}

Following previous work of ~\citet{wang2022feeding}, we assess WYWRC dataset for the 4-dimensional MRC capabilities, which are word reading, sentence reading, word understanding and sentence understanding. To adapt to classical Chinese, we make some adjustments to the metrics and establish necessary dictionaries. The main adjustment is oriented towards readability, as Chinese and English have significant differences. Instead of the Flesch-Kincaid index, which could not be suitable for Chinese, we adopt the readability index for Chinese issued by \citep{jiang2020readability}.
The final evaluation results are shown in Figure~\ref{wywrc:css}. As when the capability-speciﬁc value score increases, the difficulty of the problem that the model is understanding also increases and the correlation between sentence understanding and model performance is higher ~\cite{wang2022feeding}, we can see that the proportion of high scores in the WYWRC's sentence understanding ($v_4$) category is relatively large, thus poses a huge challenge to the pre-trained models.

In Table~\ref{tab:resultsWYWRC}, we probe the accuracy of the test set to find that there are significant differences in scores between the different types of samples. Since the number of type 8 is small and it is a really hard question requiring extra knowledge, it is not very meaningful to discuss type 8.

It can be seen that Type 2 is the most difficult of the categories with a score of 17.9. In this type of question, given a list of characteristics or actions of the protagonist in a passage, the test-taker is required to select one option that either conforms to or does not conform to the given information. Some of the questions require the test-taker to have some reasoning ability. 

Type 5 is the easiest category of problems to solve, with an average score of 66.5. This type of problem involves segmenting sentences with contextual information, similar to the PUNC task, with the difference being that the four options are confusing to each other. Nevertheless, because the problem is relatively simple and does not require much knowledge, it is easy for the models to handle.

Type 0 is the category with the largest number of samples, but its score is not high, indicating a considerable level of challenge. This type of problem involves several statements about an article, and the task is to identify the correct or incorrect option. Solving this type of problem requires not only understanding the content of the article, but also sometimes additional knowledge support and even reasoning. Therefore, this type of problem not only requires a good pre-trained model, but may also requires better fine-tuning methods and the injection of more knowledge to achieve better accuracy.

All models have a significant gap compared to human scores, mainly due to the additional knowledge and reasoning abilities of the testers. This suggests that while current natural language processing models have made significant progress in understanding and handling language, there is still a long way to go before they can approach the level of human language comprehension and reasoning.

\section{Conclusions and Future Work}
In this paper, we introduce a NLP benchmark for classical Chinese, which contains nine NLP tasks and datasets respectively to help researchers to evaluate and analyze NLP models. Also, we build a toolkit for reproduction and a leaderboard online for the community. 

The study of ancient Chinese is a highly specialized subject, so the professionalism of this benchmark may need to be further improved. On the other hand, there is a big gap between the performance of the classical Chinese models on this benchmark with other leader-boards. Better models are needed to handle more linguistic features of classical Chinese.
Furthermore, to resolve traditional and simplified character issue, traditional style tasks are meaningful to researchers. We consider it as a future work of the community.

Classical Chinese is a treasure of the entire human cultural history. We contribute this work with the hope of helping the entire community to be more prosperous. Our work will be an open, community-driven project which improves with the advancement of technology.

\section{Limitations}

In this work, we contribute an evaluation benchmark for classical Chinese NLP tasks. We did our best to create as comprehensive a well-defined task set as possible, something no one has done before. However, our work has several limitations due to lacking expertise knowledge and data.

When designing the tasks, we got a lot of inspiration from the middle school Chinese test paper. thousands of test papers are collected in order to extract data for NLP tasks. During the work process, we learn that it is difficult to extract a sufficient number of questions of a single type. The main difficulty is due to the variety of questions on the test papers and the mixture of the language of classical and modern Chinese. Finally, we create Xuci task, WYWRC task and IRC task from the test papers and related literature but failed to create solvable natural language inference tasks.

When working on some datasets which have less corpus, i.e, the Xuci task, we find it very difficult to calibrate existing samples or create new ones, resulting a small dataset size. 

Meanwhile, the category rule we followed in the GJC task is not certified by authoritative experts, so this method is not completely reliable if viewed by experts of classical Chinese.

In this work, tasks for more aspects of grammar phenomenon are lacking. It's expected that more classical Chinese experts and researchers join this work in the future to solve the above problems.

On the other hand, we lack a diagnostic dataset compared to other benchmarks. This is because similar data (NLI corpus generally) are even more difficult to retrieve. However, this benchmark works for NLP researchers even though the diagnostic dataset is missing. This issue is also expected to be solved in future work.

\section{Impact Statement}
This work aims to help in enhancing the capabilities of pre-trained models for classical Chinese language basic infrastructure. Classical Chinese is a wealth of all humanity, with a great influence worldwide. We hope to help to improve the prosperity of the classical Chinese NLP community and better mine this spiritual wealth.

Our data sources, including raw classical Chinese text and related modern text, mainly come from official and authoritative releases, so there are generally not many ethical concerns. As mentioned above, when using copyrighted texts, we have got their permission. In terms of data sets, we expect to objectively and comprehensively reflect the language characteristics of classical Chinese as much as possible, so we also try to be as general as possible when sampling, without deliberately doing content filtering. 

\paragraph{Bias and Race Concern} Classical Chinese was born in ancient times when people entered the patriarchal society, and it is inevitable to be some gender bias. Since such content is not common, we believe that there will not be many ethical issues. Additionally, one of the most important categories in ancient texts - historical books - records a large number of wars between the central government and surrounding minority ethnic groups, so there are derogatory and insulting terms among different ethnic groups. But fusions of different ethnic groups have been more often in the history books and no contents so-called racial discrimination in the current society exist.

\paragraph{Energy Cost} Practical pre-training language model requires large amounts of computation, so the cost and efﬁciency of such models should be taken into account\cite{brown2020language}. To achieve a better score on the leaderboard, many tries of pre-training may be required hence a large amount of energy consumption. So models with better efficiency are preferred for environment-friendly reasons. For instance, in this work, the DeBERTa-base model scores better than the large models which have three times more parameters. 

\section*{Acknowledgements}

We thank everyone who contributes dataset to this project. We are also grateful to the annotators and scholars who have spent much of their time and effort helping with the creation of this benchmark. This work was supported by Zhejiang Provincial Natural Science Foundation of China under Grant No.~LZ23F020009, the NSFC under Grant No.~62072399, Chinese Knowledge Center for Engineering Sciences and Technology, MoE Engineering Research Center of Digital Library. Thanks to Hangzhou AI Computing Center for the computing resource. Special thanks to the following companies and organizations: Hangzhou Xiaoniao AI Co., Ltd., Hangzhou Bolazhe Co., Ltd. Gulian (Beijing) media tech. Co., Ltd.

\bibliographystyle{acl_natbib}
\bibliography{anthology, custom}

\begin{thebibliography}{54}
\expandafter\ifx\csname natexlab\endcsname\relax\def\natexlab#1{#1}\fi

\bibitem[{Brown et~al.(2020)Brown, Mann, Ryder, Subbiah, Kaplan, Dhariwal,
  Neelakantan, Shyam, Sastry, Askell et~al.}]{brown2020language}
Tom Brown, Benjamin Mann, Nick Ryder, Melanie Subbiah, Jared~D Kaplan, Prafulla
  Dhariwal, Arvind Neelakantan, Pranav Shyam, Girish Sastry, Amanda Askell,
  et~al. 2020.
\newblock Language models are few-shot learners.
\newblock \emph{Advances in neural information processing systems},
  33:1877--1901.

\bibitem[{Chang et~al.(2021)Chang, Shiue, Yeh, and Demberg}]{chang2021time}
Ernie Chang, Yow-Ting Shiue, Hui-Syuan Yeh, and Vera Demberg. 2021.
\newblock Time-aware ancient chinese text translation and inference.
\newblock \emph{arXiv preprint arXiv:2107.03179}.

\bibitem[{Conneau and Kiela(2018)}]{conneau2018senteval}
Alexis Conneau and Douwe Kiela. 2018.
\newblock Senteval: An evaluation toolkit for universal sentence
  representations.
\newblock \emph{arXiv preprint arXiv:1803.05449}.

\bibitem[{Cui et~al.(2020{\natexlab{a}})Cui, Che, Liu, Qin, Wang, and
  Hu}]{cui2020revisiting}
Yiming Cui, Wanxiang Che, Ting Liu, Bing Qin, Shijin Wang, and Guoping Hu.
  2020{\natexlab{a}}.
\newblock Revisiting pre-trained models for chinese natural language
  processing.
\newblock \emph{arXiv preprint arXiv:2004.13922}.

\bibitem[{Cui et~al.(2021{\natexlab{a}})Cui, Che, Liu, Qin, and
  Yang}]{cui2021preCBert}
Yiming Cui, Wanxiang Che, Ting Liu, Bing Qin, and Ziqing Yang.
  2021{\natexlab{a}}.
\newblock Pre-training with whole word masking for chinese bert.
\newblock \emph{IEEE/ACM Transactions on Audio, Speech, and Language
  Processing}, 29:3504--3514.

\bibitem[{Cui et~al.(2021{\natexlab{b}})Cui, Che, Liu, Qin, and
  Yang}]{cui-etal-2021-pretrain}
Yiming Cui, Wanxiang Che, Ting Liu, Bing Qin, and Ziqing Yang.
  2021{\natexlab{b}}.
\newblock \href {https://doi.org/10.1109/TASLP.2021.3124365} {Pre-training with
  whole word masking for chinese bert}.

\bibitem[{Cui et~al.(2018)Cui, Liu, Che, Xiao, Chen, Ma, Wang, and
  Hu}]{cui2018span}
Yiming Cui, Ting Liu, Wanxiang Che, Li~Xiao, Zhipeng Chen, Wentao Ma, Shijin
  Wang, and Guoping Hu. 2018.
\newblock A span-extraction dataset for chinese machine reading comprehension.
\newblock \emph{arXiv preprint arXiv:1810.07366}.

\bibitem[{Cui et~al.(2020{\natexlab{b}})Cui, Liu, Yang, Chen, Ma, Che, Wang,
  and Hu}]{cui-etal-2020-sentence}
Yiming Cui, Ting Liu, Ziqing Yang, Zhipeng Chen, Wentao Ma, Wanxiang Che,
  Shijin Wang, and Guoping Hu. 2020{\natexlab{b}}.
\newblock \href {https://doi.org/10.18653/v1/2020.coling-main.589} {A sentence
  cloze dataset for {C}hinese machine reading comprehension}.
\newblock In \emph{Proceedings of the 28th International Conference on
  Computational Linguistics}, pages 6717--6723, Barcelona, Spain (Online).
  International Committee on Computational Linguistics.

\bibitem[{Devlin et~al.(2019)Devlin, Chang, Lee, and
  Toutanova}]{devlin-etal-2019-bert}
Jacob Devlin, Ming-Wei Chang, Kenton Lee, and Kristina Toutanova. 2019.
\newblock \href {https://doi.org/10.18653/v1/N19-1423} {{BERT}: Pre-training of
  deep bidirectional transformers for language understanding}.
\newblock In \emph{Proceedings of the 2019 Conference of the North {A}merican
  Chapter of the Association for Computational Linguistics: Human Language
  Technologies, Volume 1 (Long and Short Papers)}, pages 4171--4186,
  Minneapolis, Minnesota. Association for Computational Linguistics.

\bibitem[{Dong et~al.(2019)Dong, Yang, Wang, Wei, Liu, Wang, Gao, Zhou, and
  Hon}]{dong2019unified}
Li~Dong, Nan Yang, Wenhui Wang, Furu Wei, Xiaodong Liu, Yu~Wang, Jianfeng Gao,
  Ming Zhou, and Hsiao-Wuen Hon. 2019.
\newblock Unified language model pre-training for natural language
  understanding and generation.
\newblock \emph{Advances in Neural Information Processing Systems}, 32.

\bibitem[{Duan et~al.(2019)Duan, Wang, Wang, Ma, Cui, Wu, Wang, Liu, Huo, Hu
  et~al.}]{duan2019cjrc}
Xingyi Duan, Baoxin Wang, Ziyue Wang, Wentao Ma, Yiming Cui, Dayong Wu, Shijin
  Wang, Ting Liu, Tianxiang Huo, Zhen Hu, et~al. 2019.
\newblock Cjrc: A reliable human-annotated benchmark dataset for chinese
  judicial reading comprehension.
\newblock In \emph{China National Conference on Chinese Computational
  Linguistics}, pages 439--451. Springer.

\bibitem[{GULIAN(2020)}]{GULIAN2020}
GULIAN. 2020.
\newblock \href {http://www.gujilianhe.com/} {"gulian cup" ancient book
  document named entity recognition competition of ccl 2020}.

\bibitem[{Guzm{\'a}n et~al.(2019)Guzm{\'a}n, Chen, Ott, Pino, Lample, Koehn,
  Chaudhary, and Ranzato}]{guzman2019flores}
Francisco Guzm{\'a}n, Peng-Jen Chen, Myle Ott, Juan Pino, Guillaume Lample,
  Philipp Koehn, Vishrav Chaudhary, and Marc'Aurelio Ranzato. 2019.
\newblock The flores evaluation datasets for low-resource machine translation:
  Nepali-english and sinhala-english.
\newblock \emph{arXiv preprint arXiv:1902.01382}.

\bibitem[{He et~al.(2020)He, Liu, Gao, and Chen}]{he2020deberta}
Pengcheng He, Xiaodong Liu, Jianfeng Gao, and Weizhu Chen. 2020.
\newblock Deberta: Decoding-enhanced bert with disentangled attention.
\newblock In \emph{International Conference on Learning Representations}.

\bibitem[{Hu~Renfen(2021)}]{HU2021SentSeg}
Zhu~Yuchen Hu~Renfen, Li~Shen. 2021.
\newblock \href {http://jcip.cipsc.org.cn/CN/Y2021/V35/I4/8} {Knowledge
  representation and sentence segmentation of ancient chinese based on deep
  language models}.
\newblock \emph{JOURNAL OF CHINESE INFORMATION PROCESSING}, 35(4):8--15.

\bibitem[{Jin(2004)}]{Jin:2004}
Jishi Jin. 2004.
\newblock A brief critical summuary of the chinese language education in rok.
\newblock \emph{DongJiang Journal}, 21(1).

\bibitem[{Koichi et~al.(2022)Koichi, Christian, Tomohiko, Takumi, Naoki,
  Yoshihiro, Shingo, Shigeki, and Kazunori}]{Yasuoka:2022}
Yasuoka Koichi, Wittern Christian, Morioka Tomohiko, Ikeda Takumi, Yamazaki
  Naoki, Nikaido Yoshihiro, Suzuki Shingo, Moro Shigeki, and Fujita Kazunori.
  2022.
\newblock Designing universal dependencies for classical chinese and its
  application.
\newblock \emph{Journal of Information Processing Society of Japan},
  63(2):355--363.

\bibitem[{Lai et~al.(2017)Lai, Xie, Liu, Yang, and Hovy}]{lai2017race}
Guokun Lai, Qizhe Xie, Hanxiao Liu, Yiming Yang, and Eduard Hovy. 2017.
\newblock Race: Large-scale reading comprehension dataset from examinations.
\newblock \emph{arXiv preprint arXiv:1704.04683}.

\bibitem[{Lan et~al.(2019)Lan, Chen, Goodman, Gimpel, Sharma, and
  Soricut}]{lan2019albert}
Zhenzhong Lan, Mingda Chen, Sebastian Goodman, Kevin Gimpel, Piyush Sharma, and
  Radu Soricut. 2019.
\newblock Albert: A lite bert for self-supervised learning of language
  representations.
\newblock \emph{arXiv preprint arXiv:1909.11942}.

\bibitem[{Li et~al.(2013)Li, Gao, and Cui}]{HYFZS2013}
Guangjie Li, Xiaomei Gao, and Xiulan Cui. 2013.
\newblock \emph{汉语发展史研究}.
\newblock 黑龙江大学出版社.

\bibitem[{Li(2002)}]{Li2002GJZL}
Guoxin Li. 2002.
\newblock The development and task of chinese ancient book resources
  digitization.
\newblock \emph{Journal of Academic Libraries}, (1):21--26.

\bibitem[{Li et~al.(2021)Li, Qi, Sun, Yi, and Zhang}]{li2021CCPM}
Wenhao Li, Fanchao Qi, Maosong Sun, Xiaoyuan Yi, and Jiarui Zhang. 2021.
\newblock Ccpm: A chinese classical poetry matching dataset.
\newblock \emph{arXiv preprint arXiv:2106.01979}.

\bibitem[{Liu et~al.(1995)Liu, Cao, and Wu}]{Liu1995XUCI}
Jian Liu, Guangshun Cao, and Fuxiang Wu. 1995.
\newblock Several factors that induce lexical-grammaticalization in chinese.
\newblock \emph{Chinese language}, (3):161--169.

\bibitem[{Liu et~al.(2019)Liu, Ott, Goyal, Du, Joshi, Chen, Levy, Lewis,
  Zettlemoyer, and Stoyanov}]{liu2019roberta}
Yinhan Liu, Myle Ott, Naman Goyal, Jingfei Du, Mandar Joshi, Danqi Chen, Omer
  Levy, Mike Lewis, Luke Zettlemoyer, and Veselin Stoyanov. 2019.
\newblock Roberta: A robustly optimized bert pretraining approach.
\newblock \emph{arXiv preprint arXiv:1907.11692}.

\bibitem[{Paszke et~al.(2019)Paszke, Gross, Massa, Lerer, Bradbury, Chanan,
  Killeen, Lin, Gimelshein, Antiga, Desmaison, K\"{o}pf, Yang, DeVito, Raison,
  Tejani, Chilamkurthy, Steiner, Fang, Bai, and Chintala}]{paszke2019pytorch}
Adam Paszke, Sam Gross, Francisco Massa, Adam Lerer, James Bradbury, Gregory
  Chanan, Trevor Killeen, Zeming Lin, Natalia Gimelshein, Luca Antiga, Alban
  Desmaison, Andreas K\"{o}pf, Edward Yang, Zach DeVito, Martin Raison, Alykhan
  Tejani, Sasank Chilamkurthy, Benoit Steiner, Lu~Fang, Junjie Bai, and Soumith
  Chintala. 2019.
\newblock \emph{PyTorch: An Imperative Style, High-Performance Deep Learning
  Library}. Curran Associates Inc., Red Hook, NY, USA.

\bibitem[{Phong1 and Van2(2020)}]{Nguyen:2020}
Nguyen~Xuan Phong1 and Vu~Hong Van2. 2020.
\newblock \href
  {https://www.jnronline.com/ojs/index.php/about/article/view/577} {Taoism in
  vietnam during the northern colonial period and some notes when studying
  taoism in vietnam}.
\newblock \emph{Journal of Natural Remedies}, 21(8(1)):342--352.

\bibitem[{Qi et~al.(2022)Qi, Yang, Yi, Cheng, Liu, and
  Sun}]{qi-etal-2022-quoter}
Fanchao Qi, Yanhui Yang, Jing Yi, Zhili Cheng, Zhiyuan Liu, and Maosong Sun.
  2022.
\newblock \href {https://aclanthology.org/2022.acl-long.27} {{Q}uote{R}: A
  benchmark of quote recommendation for writing}.
\newblock In \emph{Proceedings of the 60th Annual Meeting of the Association
  for Computational Linguistics (Volume 1: Long Papers)}, pages 336--348,
  Dublin, Ireland. Association for Computational Linguistics.

\bibitem[{Qi(2022)}]{Qi2022GJZL}
Jianglei Qi. 2022.
\newblock Strategies for the development of a platform for ancient text
  knowledge service.
\newblock \emph{Chinese Editors Journal}, (2):60--65.

\bibitem[{Qiu et~al.(2020)Qiu, Sun, Xu, Shao, Dai, and Huang}]{qiu2020pre}
Xipeng Qiu, Tianxiang Sun, Yige Xu, Yunfan Shao, Ning Dai, and Xuanjing Huang.
  2020.
\newblock Pre-trained models for natural language processing: A survey.
\newblock \emph{Science China Technological Sciences}, 63(10):1872--1897.

\bibitem[{Raffel et~al.(2019)Raffel, Shazeer, Roberts, Lee, Narang, Matena,
  Zhou, Li, and Liu}]{raffel2019exploring}
Colin Raffel, Noam Shazeer, Adam Roberts, Katherine Lee, Sharan Narang, Michael
  Matena, Yanqi Zhou, Wei Li, and Peter~J Liu. 2019.
\newblock Exploring the limits of transfer learning with a unified text-to-text
  transformer.
\newblock \emph{arXiv preprint arXiv:1910.10683}.

\bibitem[{Shao et~al.(2021)Shao, Shao, Wang, Wang, and
  Gao}]{10.1145/3459637.3481964}
Yizhan Shao, Tong Shao, Minghao Wang, Peng Wang, and Jie Gao. 2021.
\newblock \href {https://doi.org/10.1145/3459637.3481964} {A sentiment and
  style controllable approach for chinese poetry generation}.
\newblock In \emph{Proceedings of the 30th ACM International Conference on
  Information \& Knowledge Management}, CIKM '21, page 4784–4788, New York,
  NY, USA. Association for Computing Machinery.

\bibitem[{Sun et~al.(2019)Sun, Yu, Yu, and Cardie}]{sun2019probing}
Kai Sun, Dian Yu, Dong Yu, and Claire Cardie. 2019.
\newblock Probing prior knowledge needed in challenging chinese machine reading
  comprehension.
\newblock \emph{arXiv preprint arXiv:1904.09679}.

\bibitem[{Wang et~al.(2019{\natexlab{a}})Wang, Pruksachatkun, Nangia, Singh,
  Michael, Hill, Levy, and Bowman}]{NEURIPS2019_4496bf24}
Alex Wang, Yada Pruksachatkun, Nikita Nangia, Amanpreet Singh, Julian Michael,
  Felix Hill, Omer Levy, and Samuel Bowman. 2019{\natexlab{a}}.
\newblock \href
  {https://proceedings.neurips.cc/paper/2019/file/4496bf24afe7fab6f046bf4923da8de6-Paper.pdf}
  {Superglue: A stickier benchmark for general-purpose language understanding
  systems}.
\newblock In \emph{Advances in Neural Information Processing Systems},
  volume~32. Curran Associates, Inc.

\bibitem[{Wang et~al.(2019{\natexlab{b}})Wang, Singh, Michael, Hill, Levy, and
  Bowman}]{wang2018glue}
Alex Wang, Amanpreet Singh, Julian Michael, Felix Hill, Omer Levy, and
  Samuel~R. Bowman. 2019{\natexlab{b}}.
\newblock \href {https://openreview.net/forum?id=rJ4km2R5t7} {{GLUE}: A
  multi-task benchmark and analysis platform for natural language
  understanding}.
\newblock In \emph{International Conference on Learning Representations}.

\bibitem[{Wang et~al.(2021)Wang, Liu, Zhu, Liu, Hu, Shen, and
  Li}]{wangdongbo:sikubert2021}
Dongbo Wang, Chang Liu, Zihe Zhu, Jiangfeng Liu, Haotian Hu, Si~Shen, and Bin
  Li. 2021.
\newblock \href
  {https://kns.cnki.net/kcms/detail/44.1306.G2.20210819.2052.008.html}
  {Sikubert 与
  sikuroberta：面向数字人文的《四库全书》预训练模型构建及应用研究}.
\newblock \emph{Library Tribune}.

\bibitem[{Wang(2004)}]{WangLi2004}
Li~Wang. 2004.
\newblock \emph{汉语史稿}.
\newblock 中华书局.

\bibitem[{Wang et~al.(2019{\natexlab{c}})Wang, Bi, Yan, Wu, Bao, Xia, Peng, and
  Si}]{wang2019structbert}
Wei Wang, Bin Bi, Ming Yan, Chen Wu, Zuyi Bao, Jiangnan Xia, Liwei Peng, and
  Luo Si. 2019{\natexlab{c}}.
\newblock Structbert: Incorporating language structures into pre-training for
  deep language understanding.
\newblock \emph{arXiv preprint arXiv:1908.04577}.

\bibitem[{Wang et~al.(2022)Wang, Liu, Xu, Long, Tang, and Wu}]{wang2022feeding}
Xiaoqiang Wang, Bang Liu, Fangli Xu, Bo~Long, Siliang Tang, and Lingfei Wu.
  2022.
\newblock Feeding what you need by understanding what you learned.
\newblock \emph{arXiv preprint arXiv:2203.02753}.

\bibitem[{Wei et~al.(2019)Wei, Ren, Li, Huang, Liao, Wang, Lin, Jiang, Chen,
  and Liu}]{wei2019nezha}
Junqiu Wei, Xiaozhe Ren, Xiaoguang Li, Wenyong Huang, Yi~Liao, Yasheng Wang,
  Jiashu Lin, Xin Jiang, Xiao Chen, and Qun Liu. 2019.
\newblock Nezha: Neural contextualized representation for chinese language
  understanding.
\newblock \emph{arXiv preprint arXiv:1909.00204}.

\bibitem[{Wolf et~al.(2020)Wolf, Debut, Sanh, Chaumond, Delangue, Moi, Cistac,
  Rault, Louf, Funtowicz et~al.}]{wolf2020transformers}
Thomas Wolf, Lysandre Debut, Victor Sanh, Julien Chaumond, Clement Delangue,
  Anthony Moi, Pierric Cistac, Tim Rault, R{\'e}mi Louf, Morgan Funtowicz,
  et~al. 2020.
\newblock Transformers: State-of-the-art natural language processing.
\newblock In \emph{Proceedings of the 2020 conference on empirical methods in
  natural language processing: system demonstrations}, pages 38--45.

\bibitem[{Xiang et~al.(2021)Xiang, Yang, Li, Warstadt, and
  Kann}]{xiang-etal-2021-climp}
Beilei Xiang, Changbing Yang, Yu~Li, Alex Warstadt, and Katharina Kann. 2021.
\newblock \href {https://doi.org/10.18653/v1/2021.eacl-main.242} {{CL}i{MP}: A
  benchmark for {C}hinese language model evaluation}.
\newblock In \emph{Proceedings of the 16th Conference of the European Chapter
  of the Association for Computational Linguistics: Main Volume}, pages
  2784--2790, Online. Association for Computational Linguistics.

\bibitem[{Xu et~al.(2020)Xu, Hu, Zhang, Li, Cao, Li, Xu, Sun, Yu, Yu, Tian,
  Dong, Liu, Shi, Cui, Li, Zeng, Wang, Xie, Li, Patterson, Tian, Zhang, Zhou,
  Liu, Zhao, Zhao, Yue, Zhang, Yang, Richardson, and Lan}]{xu2020clue}
Liang Xu, Hai Hu, Xuanwei Zhang, Lu~Li, Chenjie Cao, Yudong Li, Yechen Xu, Kai
  Sun, Dian Yu, Cong Yu, Yin Tian, Qianqian Dong, Weitang Liu, Bo~Shi, Yiming
  Cui, Junyi Li, Jun Zeng, Rongzhao Wang, Weijian Xie, Yanting Li, Yina
  Patterson, Zuoyu Tian, Yiwen Zhang, He~Zhou, Shaoweihua Liu, Zhe Zhao, Qipeng
  Zhao, Cong Yue, Xinrui Zhang, Zhengliang Yang, Kyle Richardson, and Zhenzhong
  Lan. 2020.
\newblock \href {https://doi.org/10.18653/v1/2020.coling-main.419} {{CLUE}: A
  {C}hinese language understanding evaluation benchmark}.
\newblock In \emph{Proceedings of the 28th International Conference on
  Computational Linguistics}, pages 4762--4772, Barcelona, Spain (Online).
  International Committee on Computational Linguistics.

\bibitem[{Xu(1995)}]{Xu:95}
Qiuhan Xu. 1995.
\newblock \href {http://lib.cqvip.com/Qikan/Article/Detail?id=1002530270}
  {Chinese characters in japan.}
\newblock \emph{中国文化研究}, pages 135--139+6.

\bibitem[{Yang et~al.(2019)Yang, Dai, Yang, Carbonell, Salakhutdinov, and
  Le}]{yang2019xlnet}
Zhilin Yang, Zihang Dai, Yiming Yang, Jaime Carbonell, Russ~R Salakhutdinov,
  and Quoc~V Le. 2019.
\newblock Xlnet: Generalized autoregressive pretraining for language
  understanding.
\newblock \emph{Advances in neural information processing systems}, 32.

\bibitem[{Yang et~al.(2021)Yang, Chen, and Chen}]{yang2021guwenunilm}
Zinong Yang, Ke-jia Chen, and Jingqiang Chen. 2021.
\newblock Guwen-unilm: Machine translation between ancient and modern chinese
  based on pre-trained models.
\newblock In \emph{CCF International Conference on Natural Language Processing
  and Chinese Computing}, pages 116--128. Springer.

\bibitem[{Yao et~al.(2021)Yao, Dong, Guan, Cao, Zhang, Xiao, Wang, Qi, Bao, Nie
  et~al.}]{yao2021cuge}
Yuan Yao, Qingxiu Dong, Jian Guan, Boxi Cao, Zhengyan Zhang, Chaojun Xiao,
  Xiaozhi Wang, Fanchao Qi, Junwei Bao, Jinran Nie, et~al. 2021.
\newblock Cuge: A chinese language understanding and generation evaluation
  benchmark.
\newblock \emph{arXiv preprint arXiv:2112.13610}.

\bibitem[{Ye and Tian(2013)}]{Ye:2013}
Shaofei Ye and Zhiyong Tian. 2013.
\newblock On the origins of vietnamese ancient history.
\newblock \emph{Southeast Asian and South Asian Studies}, (2):83--89.

\bibitem[{Yin et~al.(2018)Yin, Fang, and Shen}]{Yin:2018}
Xiaolin Yin, Ming Fang, and Wenfan Shen, editors. 2018.
\newblock \emph{中华传世藏书}.
\newblock 浙江人民出版社有限公司.

\bibitem[{Yu et~al.(2021)Yu, Wei, and Zhang}]{Yu2021Seg}
Jingsong Yu, Yi~Wei, and Yongwei Zhang. 2021.
\newblock \href {http://jcip.cipsc.org.cn/CN/Y2019/V33/I11/57} {Automatic
  ancient chinese texts segmentation based on bert}.
\newblock \emph{JOURNAL OF CHINESE INFORMATION PROCESSING}, 33(11):57--63.

\bibitem[{Yuying(2020)}]{jiang2020readability}
JIANG Yuying. 2020.
\newblock \href {https://doi.org/10.19360/j.cnki.11-3303/g4.2020.12.005} {A
  study on the readability of reading test texts in chinese proficiency
  test（hsk）}.
\newblock 12.

\bibitem[{Zhang et~al.(2021)Zhang, Chen, Bi, Liang, Li, Shang, Yin, Tan, Xu,
  Huang et~al.}]{zhang2021cblue}
Ningyu Zhang, Mosha Chen, Zhen Bi, Xiaozhuan Liang, Lei Li, Xin Shang, Kangping
  Yin, Chuanqi Tan, Jian Xu, Fei Huang, et~al. 2021.
\newblock Cblue: A chinese biomedical language understanding evaluation
  benchmark.
\newblock \emph{arXiv preprint arXiv:2106.08087}.

\bibitem[{Zheng et~al.(2019)Zheng, Huang, and Sun}]{zheng2019chid}
Chujie Zheng, Minlie Huang, and Aixin Sun. 2019.
\newblock Chid: A large-scale chinese idiom dataset for cloze test.
\newblock \emph{arXiv preprint arXiv:1906.01265}.

\bibitem[{Zhou(2009)}]{Zhou:2009}
Bin Zhou. 2009.
\newblock \href {http://lib.cqvip.com/Qikan/Article/Detail?id=31846634} {A
  comprehensive study on the history presented in a series of biographies
  written in chinese in japan}.
\newblock \emph{Journal of Historiography}, pages 98--104.

\bibitem[{Zinin and Xu(2020)}]{zinin-xu-2020-corpus}
Sergey Zinin and Yang Xu. 2020.
\newblock \href {https://aclanthology.org/2020.lrec-1.98} {Corpus of {C}hinese
  dynastic histories: Gender analysis over two millennia}.
\newblock In \emph{Proceedings of the 12th Language Resources and Evaluation
  Conference}, pages 785--793, Marseille, France. European Language Resources
  Association.

\end{thebibliography}
\clearpage
\newpage
\appendix
\newcommand{\hei}{\CJKfamily{SimHei}}
\newcommand{\tabincell}[2]{\begin{tabular}{@{}#1@{}}#2\end{tabular}}  

\section{Data Details and Annotation}
In this section, we use a "[SEP]" mark to denote separation between two parts of a sample. And we try to translate the classical sentence to English to make it easier to understand. It should be noted that the English translations of the following texts are machine-translated, which are not very accurate.
\label{app:data}
\subsection{WYWRC}
\label{det:wywrc}
\subsubsection{Details}
As previously mentioned, reading comprehension is a crucial aspect of classical Chinese learning, and is tested annually in the college entrance exam.
With the assistance of middle school teachers, we have collected thousands of examples from examination papers. 
This dataset is in JSON format. Statistics are show in Figure~\ref{xuci:pie} Figure~\ref{xuci:hist} and Table~\ref{xuci:Stat}. All the samples are separated into 10 types according to the difference between the questions which are shown in Table~\ref{tab:wywrcType}. Furthermore, as shown in Figure~\ref{wywrc:css}, we carry out a competency assessment of MRC capabilities to probe the challenge of this task in many fine-grained metrics.

\begin{figure}[!t]
	\centering

  \includegraphics[scale=0.4]{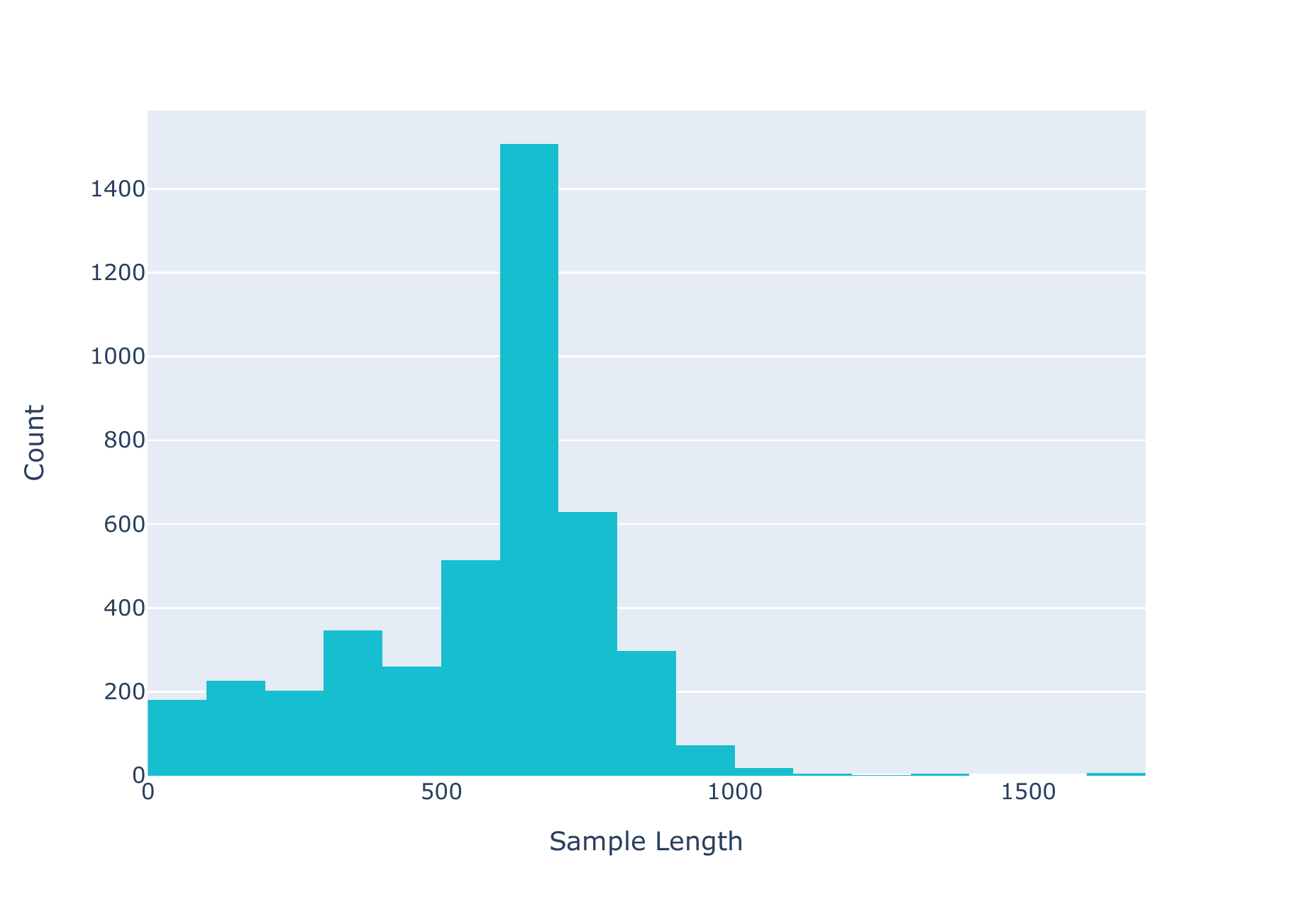}
  \caption{Statistic of article Length of WYWRC dataset.}
  \label{wywrc:article}
\end{figure}

\begin{figure}[!t]
	\centering

  \includegraphics[scale=0.4]{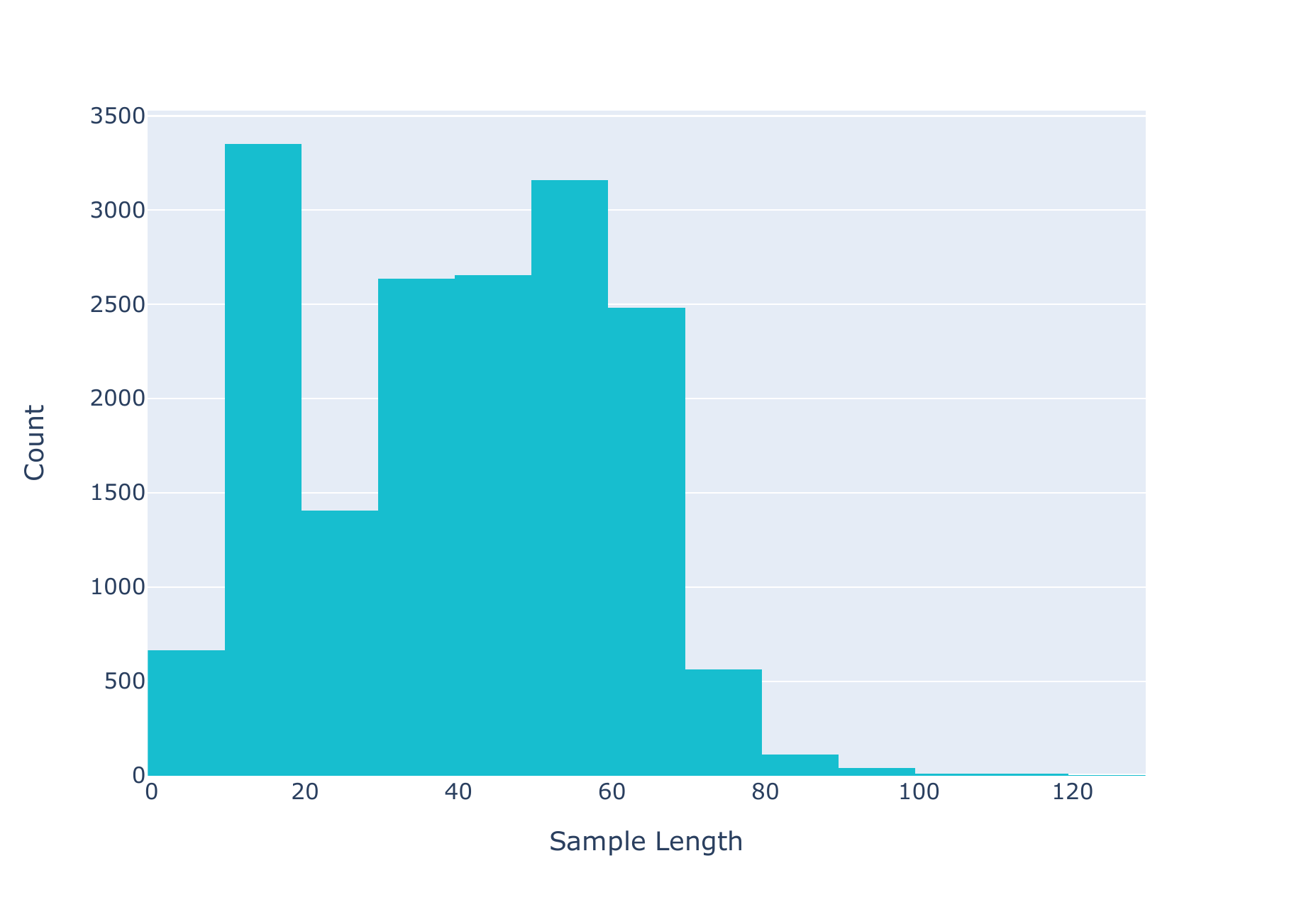}
  \caption{Statistic of option Length of WYWRC dataset.}
  \label{wywrc:option}
\end{figure}

\begin{figure}[!t]
\centering
  \includegraphics[scale=0.4]{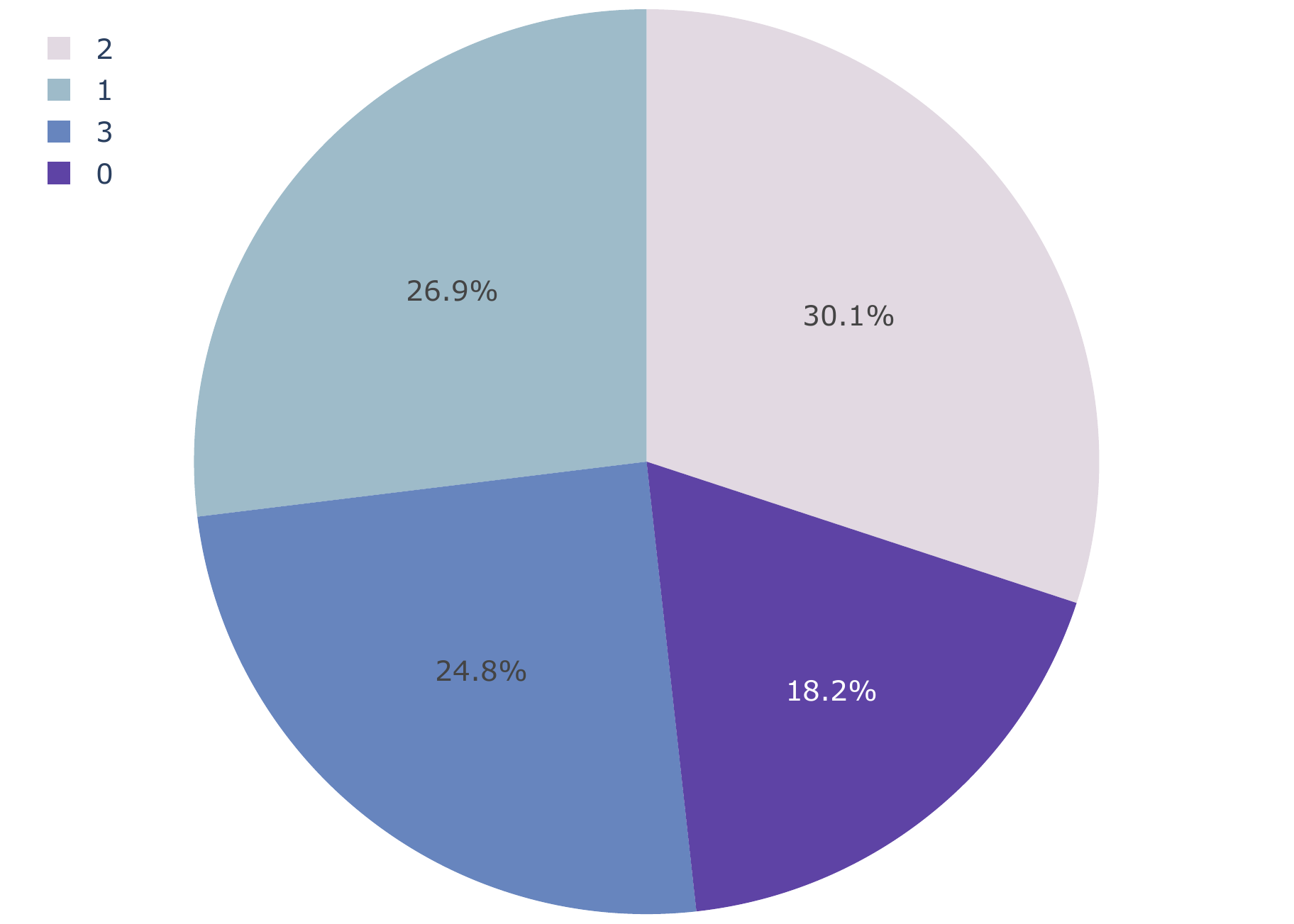}
  \caption{Statistics on the proportion of labels in the WYWRC dataset. }
  \label{wywrc:label}
\end{figure}

\begin{figure}[!t]
\centering
  \includegraphics[scale=0.4]{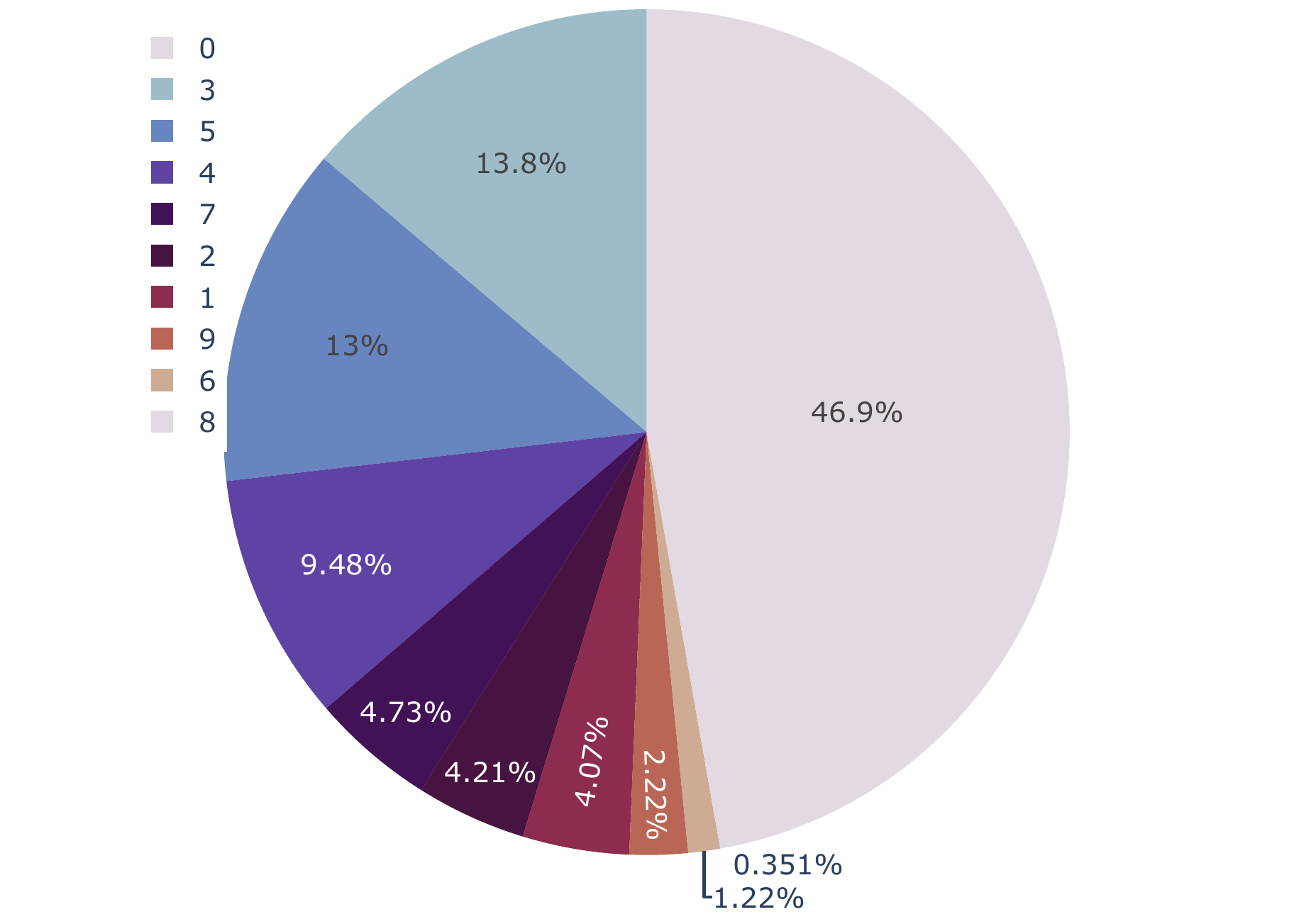}
  \caption{Statistics on the proportion of every type in the WYWRC dataset. }
  \label{wywrc:type}
\end{figure}

\begin{table*}[]
\begin{adjustbox}{width=1\textwidth}
\begin{tabular}{cllcl}
\hline \hline
\multicolumn{1}{c}{\textbf{Type}} & \multicolumn{1}{c}{\textbf{Paragraph}} & \multicolumn{1}{c|}{\textbf{Question}} & \multicolumn{1}{c}{\textbf{Answer}} & \multicolumn{1}{c}{\textbf{Options}}                \\ \hline
0   & \makecell[l]{房馆，字次律，河南人。\\琯少好学，风度沈整，以荫\\补弘文生。......\\Fang Guan, styled Cilu, was born \\in Henan. Guan Shao is eager \\to learn, has a calm demeanor, \\and supplements Hongwensheng \\with shade.}     
    & \makecell[l]{下列对原文有关内\\容的概括和分析，\\不正确的一项是\\Which of the following summary \\and analysis of the relevant content\\ of the original text is incorrect? } & 1                                   
    & \makecell[l]{0 房琯文采出众，闻名当时。开元中作《封禅书》受到宰相张说的重视；\\天宝十五年，投奔肃宗，奏事深得肃宗赏识。\\Fang Guan was famous for his outstanding literary talent at that time. Kaiyuan Zhongzuo's \\Fengchan Shu was valued by Prime Minister Zhang Shuo; In the 15th year of Tianbao's \\reign, he went to Suzong and was deeply appreciated by Suzong.\\1 房琯颇富才华，深受器重，他精于设计安排，被任命总管郦山规划；\\他参决机务，虽有较大失误，仍深受信任。\\Fang Guan was very talented and highly valued. He was good at design and arrangement,\\ and was appointed as the general manager of Li Shan's planning; He participated \\in the maintenance of the aircraft, and although he made major mistakes, he was\\ still deeply trusted.\\2 ......\\3 ......} \\ \hline
1  & \makecell[l]{鱼，我所欲也，熊掌，亦我所欲也，\\二者不可得兼，舍鱼而取熊掌者也。\\生，亦我所欲也，义，亦我所欲也，......\\Fish is what I want, bear's \\paw is also what I want. \\You can't have both. Life \\is what I want, righteousness \\is what I want, ......}  
& \makecell[l]{下列词意义相同的一项是?\\Which of the following words \\have the same meaning?}                                   
& 1                                    
& \makecell[l]{0 而|舍生而取义者也|予独爱莲之出淤泥而不染\\Er | those who give up their lives to take righteousness also \\| give the only lotus out of the mud without contamination.\\1 于|如使人之所欲莫甚于生|皆以美于徐公\\Yu | make people want more than they want to \\live | to be more beautiful than Mr. Xu\\2......\\3......}                                                   \\ \hline
2 & \makecell[l]{陈谨斋讳志鋐，字纯候。休宁有陈村，\\在县治西南山谷之间，俗尚淳朴，\\陈氏世居之。谨斋之曾祖仁琦，以孝\\悌称，......\\Chen Qingzhai tabooed his ambition and\\ wrote the word "pure waiting". There is a \\Chen village in Xiuning, which is located \\in the southwest valley of the county. \\It is simple and unsophisticated, and the \\Chen family has lived there for generations. \\Renqi, the great ancestor of Jinzhai, was \\called filial piety,}                                    
& \makecell[l]{下列各语句中，表现陈谨斋\\有“长者”之风的是?\\Among the following sentences, \\which one shows that Chen Jinzhai\\ has the style of "elderly"?}                       
& 0   
& \makecell[l]{0 守其家法尤谨\\be careful to keep the family rules\\1 所居货尝大利矣，而辄舍去之\\The goods people live in are very profitable, but they often give them away.\\2 ......\\3 ......}                                                    \\ \hline
3                                 
& \makecell[l]{童华，字心朴，浙江山阴人。雍正初，\\为知县。时方修律例，大学士朱轼荐其才，\\世宗召见，命察赈直隶。......\\Tong Hua, whose style name is Xinpu, is from \\Shanyin County in Zhejiang. During the early \\Qing Dynasty, he served as a county magistrate. \\At the time, Minister of Justice Zhu Shuo \\recommended Tong Hua's talent to Emperor\\Shizong, who summoned Tong Hua and \\appointed him as an inspector of Zhili. } 
& \makecell[l]{对下列句子中词的解释，\\不正确的一项是\\Which of the following explanations of \\the words in the sentences is incorrect?} 
& 1
& \makecell[l]{0 问滦河形势，华对甚晰。|形势：河流状况。\\When asked about the situation in Luanhe River, Hua was very clear \\about it. | Situation: The state of the river.\\1 以前在平山发粟事。|以前：从前，以往。\\In the past, millet incidents occurred in Pingshan. |Before: Once upon a time, in the past.\\2 ......\\3 ......}  \\ \hline
4  
& \makecell[l]{来护儿，字崇善，未识而孤，养于世母\\吴氏吴氏提携鞠养，甚有慈训，幼儿卓荦；\\初读《诗》，舍书叹曰：大丈夫在世，\\Lai Huer, whose character is Chongshan,\\ lonely without knowing it, was raised \\by his mother, Wu Shi, Wu Shi, and \\brought him up.}   
& \makecell[l]{下列对词语相关内容的解说，\\不正确的一项是\\Which of the following explanations \\about the content of the words \\is incorrect?} 
& 2 
& \makecell[l]{0 古代男子有名有字，名是出生后不久父亲起的，字是二十岁举行冠礼后才起的。\\In ancient times, a man had a name and a character. The name was given by his father\\ shortly after birth, and the character was not given until after the crowning \\ceremony at the age of twenty.\\1 谥号是古代帝王、大臣等死后，据其生平事迹评定的称号，如武帝、哀帝，炀帝。\\A posthumous title is a title given to an emperor, minister, or other high-ranking figure \\after their death, based on their deeds and accomplishments during their lifetime. Examples \\include Wu Di, Ai Di, and Yang Di.\\2 ......\\3 ......} \\ \hline
5 
& \makecell[l]{项羽在垓下，欲攻沛公。沛公从\\百余骑因项伯面见项羽，谢无有闭关事。项羽\\既飨军士......\\At Gai Xia, Xiang Yu intended to \\attack Peng Gong. Peng Gong, accompanied\\ by over one hundred riders, \\met with Xiang Yu face to face and thanked \\him for not closing the gates. Xiang Yu,\\ having satisfied the soldiers, then left.}            
& \makecell[l]{下列断句，正确的一项是\\Which of the following is \\the correct sentence?} 
& 1  
& \makecell[l]{0 项羽既飨军士中酒|亚父谋欲杀沛公|令项庄找剑舞坐中|欲击沛公|项伯常屏藏之\\1 项羽既飨军士|中酒|亚父谋欲杀|沛公令项庄找剑舞坐中|欲击沛公|项伯常屏藏之 \\2 ...... \\3 ......} \\ \hline
6 
& \makecell[l]{潭中鱼可百许头，皆若空游无所依，日光下澈，\\影布石上。佁然不动，俶尔远逝，往来翕忽，......\\There are numerous fish in the pool, all \\swimming aimlessly, as if there is nothing \\for them to rely on. Under the sunlight, \\they lie motionless on the surface of the stones, \\their shadows spreading out beneath them. }  
& \makecell[l]{下列句子中“而”字的用法\\与“潭西南而望”\\相同的一项是\\Which of the following sentences \\uses the word "而" in the same \\way as "潭西南而望"?} 
& 3  
& \makecell[l]{0 温故而知新\\To understand the new, one must study the old.\\1 ......\\2 ......\\3 隶而从者\\those who are subservient and follow}                                                   \\ \hline
7 
& \makecell[l]{青枥林深亦有人，一渠流水数家分。\\山当日午回峰影，草带泥痕过鹿群。......\\There are also people in the deep green \\banyan groves. A stream flows through, \\separating several homes. The shadow of \\the mountain returns to the peak at \\noon, the grass and mud tracks passing \\through the deer herd.} 
& \makecell[l]{下列对诗歌内容理解\\不正确的一项是\\Which of the following is an \\incorrect understanding of the content \\of the poem?} 
& 3   
& \makecell[l]{0 诗歌描写了诗人山行时在村里村外的所见所闻。\\The poem describes the sights and sounds the poet experiences while traveling through \\the countryside.\\1 ......\\2 ......\\3 颈联描写了烘茶的过程与抽丝的声音，展现出农事繁忙的景象。\\The description of the process of drying tea and the sound of the silkworm cocoons \\in the neckband poem depicts the busy scene of farm work.}  \\ \hline
8 
& \makecell[l]{四愁诗，张衡。一思曰，我所思兮在太山，\\欲往从之梁父艰。侧身东望涕沾翰。\\The "Four Worries", by Zhang Heng. \\As I think, my thoughts are on Mount Tai. \\I want to go there but the journey is \\difficult, as I am told by Liang Fu. \\I turn my head and look eastward, \\my writing brush dampened with tears.} 
& \makecell[l]{下列句子中，不含通假字的一项是?\\Which of the following sentences \\does not contain a homophone?} 
& 2                                    
& \makecell[l]{0 阴知奸党名姓，一时收禽\\Knowing the names and identities of the treacherous officials, I temporarily \\arrested them.\\1 ......\\2 衡少善属文\\Heng was skilled in literature.\\ 3 ......}                                                     \\ \hline
9  
& \makecell[l]{莫能名斋记，杨简。四明杨简，得屋于宝莲\\山之巅。简思所以名之，东望大江，巨涛际天，\\Mo Neng Ming Zhai Ji, Yang Jian. \\Yang Jian from Siming obtained a house on the \\peak of Baolian Mountain. Jian Si, therefore, \\named it "Looking East at the Great River, \\with Huge Tides at the Horizon."}  
& \makecell[l]{对“要不可谓真识江山”的句\\子理解正确的一项是\\Which of the following is a correct \\interpretation of the sentence \\"要不可谓真识江山"?} 
& 1                                   
& \makecell[l]{一定不能说真正认识了江山。\\We must not say that we have truly come to know the beauty of the land.\\关键是不能说真正懂得了江山。\\The key is not to think that you truly understand the rivers and mountains.}     \\ \hline \hline
\end{tabular}

\end{adjustbox}

\caption{Samples of WYWRC dataset. The samples are divided into ten categories, which are featured by the question and options.}
\label{tab:wywrcType}
\end{table*}

\begin{table*}
    \centering

    \begin{threeparttable}
      \begin{tabular}{@{}lcccccccccc@{}}
        \toprule
        \textbf{Model} & \textbf{0} & \textbf{1} & \textbf{2} & \textbf{3} & \textbf{4}      & \textbf{5} & \textbf{6} & \textbf{7} & \textbf{8} & \textbf{9}  \\ \midrule
        Human             & 85.5 & 80.4 & 77.4 & 90.5 & 80.3 & 85.8 & 70.6 & 76.7 & 50.0 & 88.6 \\ \hline
        DeBERTa-base      & 36.8 & 55.6 & 22.2 & 47.5 & 43.9 & 78.6 & 50.0 & 47.6 & 50.0 & 30.0 \\ \hline
        guwenbert-base    & 30.3 & 16.7 & 16.7 & 25.4 & 24.4 & 28.6 & 16.7 & 38.1 & 50.0 & 30.0 \\
        guwenbert-large   & 36.8 & 50.0 & 16.7 & 47.5 & 43.9 & 78.6 & 50.0 & 42.9 & 0.0  & 40.0 \\
        guwenbert-base-fs & 32.3 & 38.9 & 22.2 & 44.1 & 43.9 & 76.8 & 50.0 & 42.9 & 0.0  & 70.0 \\\hline
        roberta-CCBC      & 33.3 & 44.4 & 16.7 & 45.8 & 39.0 & 67.9 & 50.0 & 42.9 & 0.0  & 50.0 \\
        roberta-CCLC      & 36.8 & 50.0 & 16.7 & 47.5 & 43.9 & 78.6 & 50.0 & 42.9 & 0.0  & 40.0 \\ \hline
        SikuBERT          & 38.8 & 61.1 & 22.2 & 44.1 & 31.7 & 73.2 & 50.0 & 47.6 & 0.0  & 40.0 \\
        SikuRoBERTa       & 36.3 & 44.4 & 16.7 & 45.8 & 36.6 & 66.1 & 16.7 & 42.9 & 0.0  & 40.0 \\\hline
        Roberta-wwm-ext   & 37.8 & 50.0 & 11.1 & 52.5 & 39.0 & 50.0 & 33.3 & 57.1 & 0.0  & 60.0 \\ \hline
        \textbf{Average}  & 35.5 & 45.7 & 17.9 & 44.5 & 38.5 & 66.5 & 40.7 & 44.9 & 11.1 & 44.4 \\ \bottomrule
        \end{tabular}
        \end{threeparttable}
    \caption{Scores (accuracy) of every type of WYWRC dataset. The average score does not contain human performance.}
    \label{tab:resultsWYWRC}
    \end{table*}

\subsubsection{Annotation}
\begin{itemize}
\item Prepossess test papers, including OCR, layout parser and then copy all reading comprehension problems;
\item Annotate, including filter the sub-problems for which suitable for machine learning and correct misspelling, etc.;
\item Cross double-check between annotators;
\item Final review by experts.
\end{itemize}

\subsection{PUNC}
\label{det:punc}
\subsubsection{Details}
This task is designed for text punctuation. Since there are not any punctuation marks in traditional Chinese literature, discriminatory of sentence punctuation is important for reading ancient books. Even though ancient Chinese researchers have made great efforts in the proofreading and sorting out of ancient books, there are still a large number of ancient books without punctuation waiting to be solved \cite{Qi2022GJZL, Li2002GJZL}. So that punctuation task is useful for classical Chinese researchers. Therefore, all related works evaluate their models mainly on this task.

To make sure the time distribution of the corpus as uniform as possible, we select history books as source data for this task including 二十四史 (the Twenty-Four Histories), 春秋 (The Spring and Autumn Annals), 战国策 (Strategies of the Warring States Period) and so on. The corpus contains historical books from the Zhou Dynasty to the Republic of China, which cover nearly three thousand years (1046 BC to 1927). 
All of the books are concatenated and shuffled by paragraph, then sampled by a reasonable rate and finally split into task datasets. 

\begin{figure}[!t]

  \includegraphics[scale=0.4]{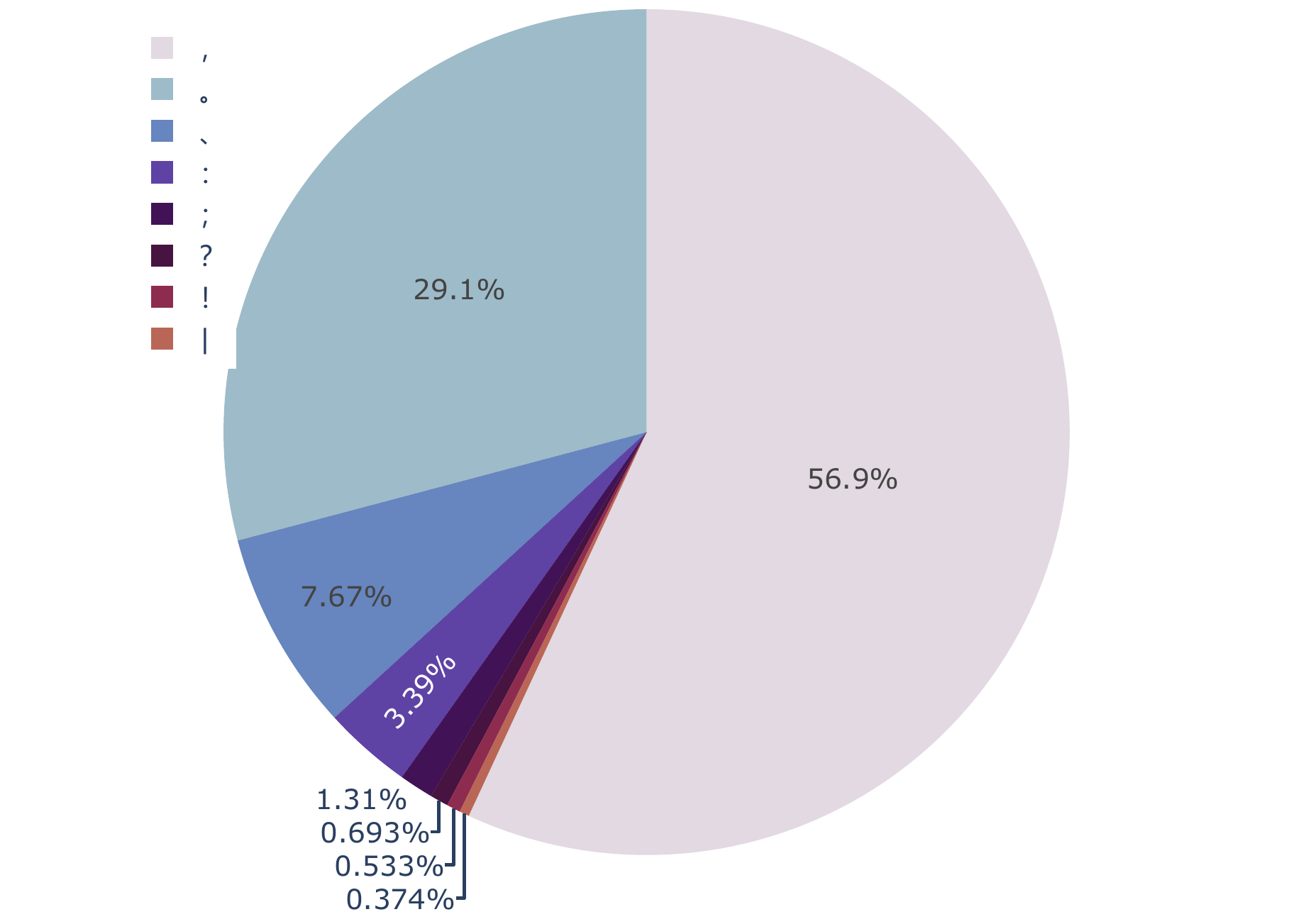}
  \caption{Percentage of punctuation marks to be predicted of PUNC dataset.}
  \label{punc:pie}
\end{figure}

\begin{figure}[ht]
  \centering
  \includegraphics[scale=0.4]{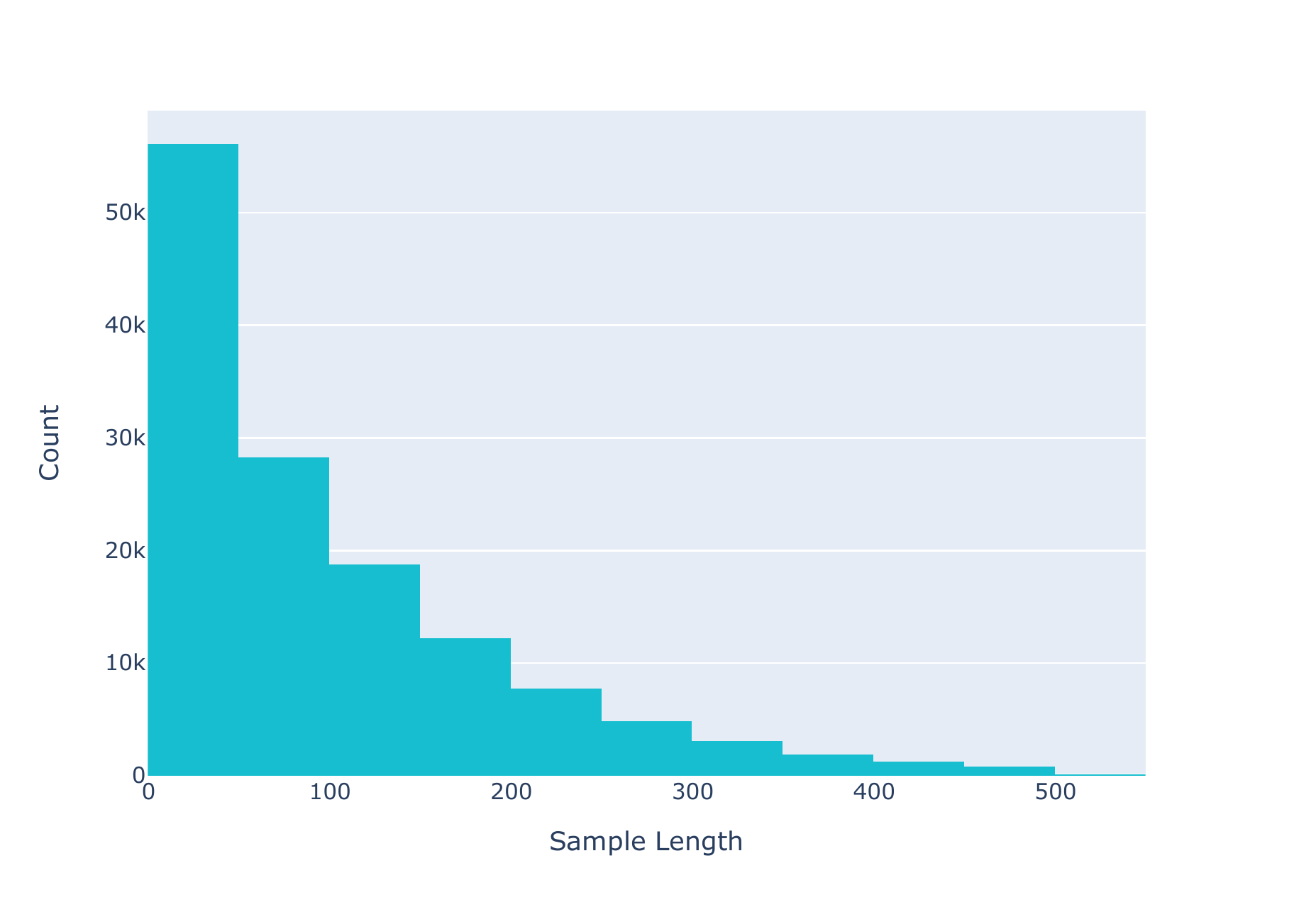}
  \caption{Statistic of Sample Length of PUNC dataset.}
  \label{punc:hist}
\end{figure}

This dataset is in sequence pair TSV format. Every sample is a pair of source text and label sequence as shown following. We choose eight punctuation marks as prediction target in this dataset. Statistics are show in Figure~\ref{punc:pie} Figure~\ref{punc:hist} and Table~\ref{punc:Stat}.

\begin{quote}
  壬戌诏定科举流寓人名额蒙古色目南人各十五名汉人二十名\\
  OO，OOOOOOOO，O、O、OOOOO，\\
  OOOO。\\
  On the 24th, an imperial edict was issued to establish the quota of expatriates in the imperial examination, 15 for Mongolians, 15 for colored-eyes and 20 for Han Chinese.\\
  
  谢肇淛《北河纪》八卷《纪余》四卷除坛西郊坎其击鼓百灵至止结璘作主\\
  OOOOOOOOOO，OOOOO|OOO，OOO\\
  。OOO，OOO。\\
  Xie Zhaozhe wrote eight volumes of Beihe Ji and four volumes of Ji Yu. At the altar in the western suburbs, playing drums, hundreds of gods stopped here and became the leader of the alliance.
\end{quote}

\begin{table}[]
  \centering

  \begin{tabular}{l|l}
  \hline
  Total Samples        & 135156 \\ \hline
  Mean Sample Length & 100    \\ \hline
  Min Sample Length  & 5      \\ \hline
  Max Sample Length  & 510    \\ \hline
  \end{tabular}
  \caption{Statistic of Sample Length of PUNC dataset.}
  \label{punc:Stat}

  \end{table}

\subsubsection{Annotation}
\begin{itemize}
    \item Extract corpora from the Twenty-Four Histories and some other history books;
    \item Filter out low quality samples, for example, too short or punctuation too few;
    \item Sample from the result corpora with a specific ratio;
    \item Annotator check;
    \item Final review by experts.
\end{itemize}











\subsection{GLNER}
\label{det:glner}
\subsubsection{Details}
\begin{figure}[!t]
	\centering
  \includegraphics[scale=0.4]{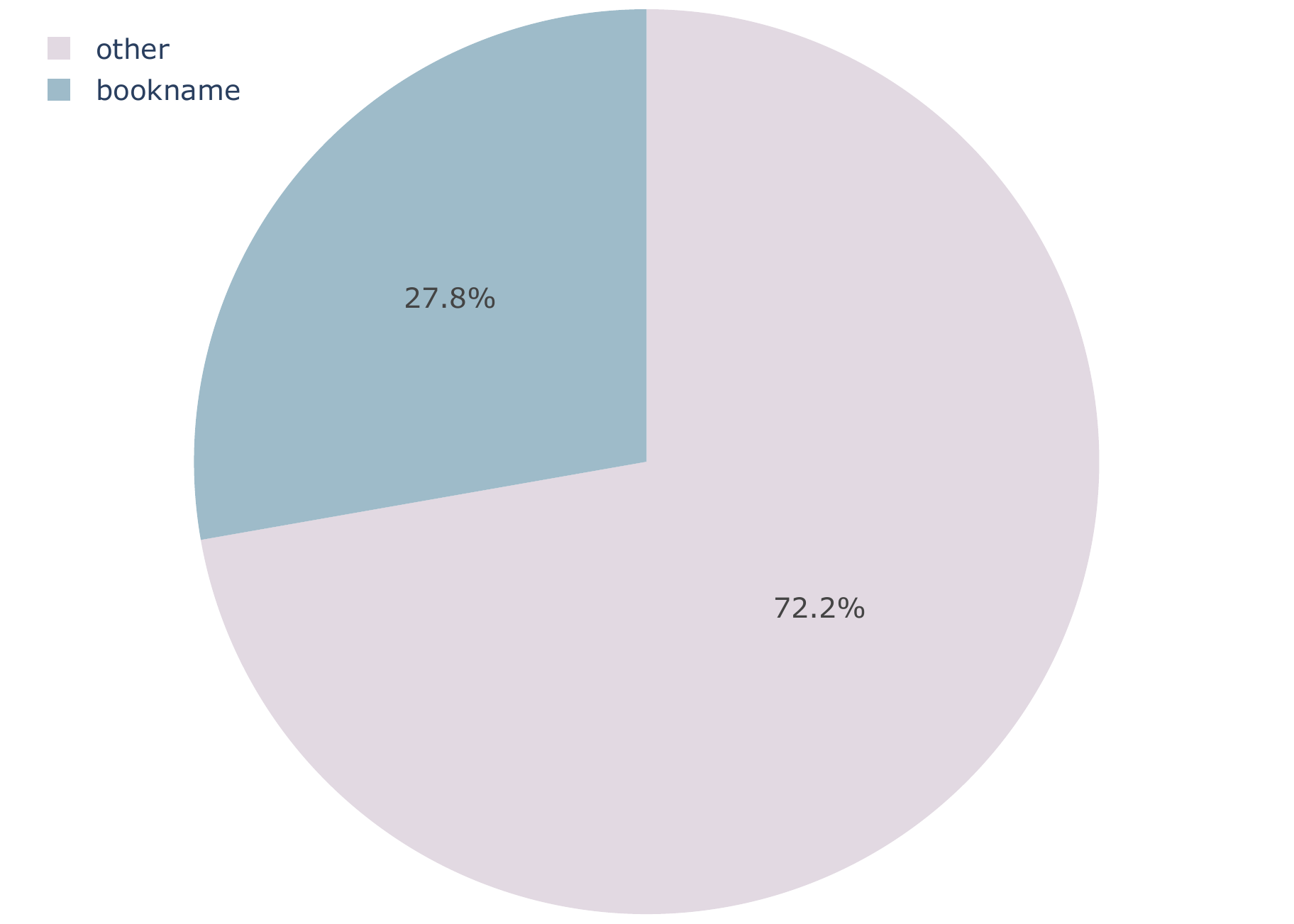}
  \caption{Percentage of labels to be predicted of GLNER dataset.}
  \label{glner:pie}
\end{figure}

\begin{figure}[!t]
	\centering

  \includegraphics[scale=0.4]{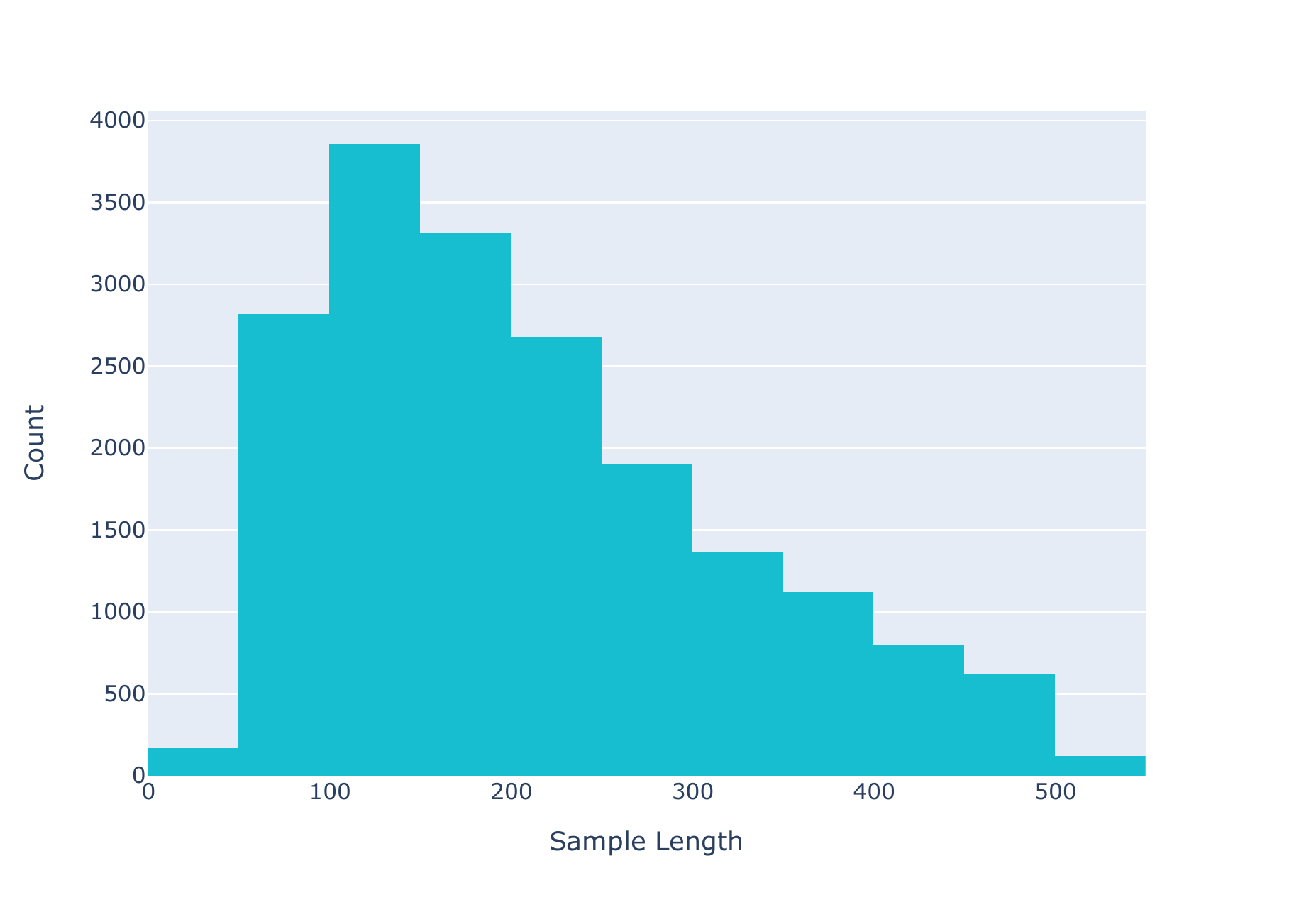}
  \caption{Statistic of Sample Length of GLNER dataset.}
  \label{glner:hist}
\end{figure}

This dataset is in JSON format as shown below. Every sample consists two keys which are "text" and "label", and every label is represented as start index, end index and category style. Statistics are show in Figure~\ref{glner:pie} Figure~\ref{glner:hist} and Table~\ref{glner:Stat}.

\begin{quote}
  \{ \\
    "text": "谢绛　三月戊戌，知礼仪院、兵部员外、知制诰谢绛知邓州。十一月己酉，卒。欧文。长编：绛按召信臣故迹，距城三里，壅湍水，注钳庐陂，溉田，请复修之。可，罢州人岁役。", \\
  "label": {[}{[}0, 2, "other"{]}, {[}21, 23, "other"{]}, {[}24, 26, "other"{]},  {[}35, 36, "other"{]}, {[}38, 40, "bookname"{]}, {[}41, 42, "other"{]}, {[}43, 46, "other"{]}, {[}59, 62, "other"{]}{]} \\
  \} \\

  \{ \\
  "text": "六月己未，郑居中等上哲宗御集。壬戌，景灵宫建禧祖殿室。复广、惠、康、贺州旧铸夹锡钱监。辛未，湖南路提点刑狱陈义夫奏邵阳县猺贼平。", \\
  "label": {[}{[}5, 8, "other"{]}, {[}10, 14, "bookname"{]}, {[}18, 21, "other"{]}, {[}22, 24, "other"{]}, {[}28, 29, "other"{]}, {[}30, 31, "other"{]}, {[}32, 33, "other"{]}, {[}34, 36, "other"{]}, {[}46, 49, "other"{]}, {[}53, 56, "other"{]}, {[}57, 61, "other"{]}{]} \\
  \}
\end{quote}

\begin{table}[]
  \centering

  \begin{tabular}{ll}
  \hline
  Total Samples        & 18762 \\ \hline
  Mean Sample Length & 210    \\ \hline
  Min Sample Length  & 28      \\ \hline
  Max Sample Length  & 510    \\ \hline
  \end{tabular}

  \caption{Statistic of Sample Length of GLNER dataset.}
  \label{glner:Stat}

  \end{table}

\subsection{TLC}
\label{det:tlc}
\subsubsection{Details}
\begin{figure}[!t]
	\centering

  \includegraphics[scale=0.4]{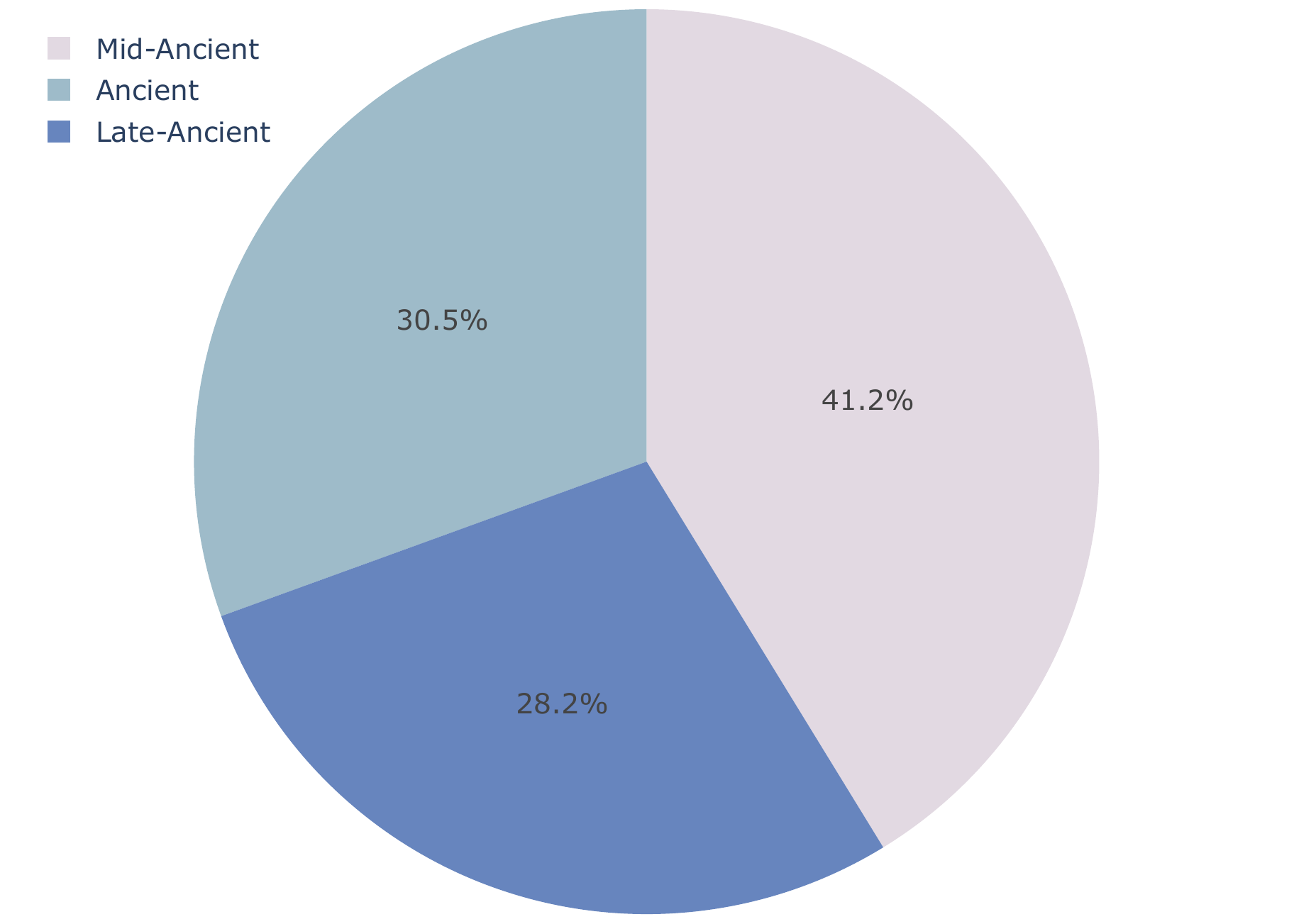}
  \caption{Percentage of Period labels.}
  \label{tlc_label_p:pie}
\end{figure}

\begin{figure}[!t]
	\centering

  \includegraphics[scale=0.4]{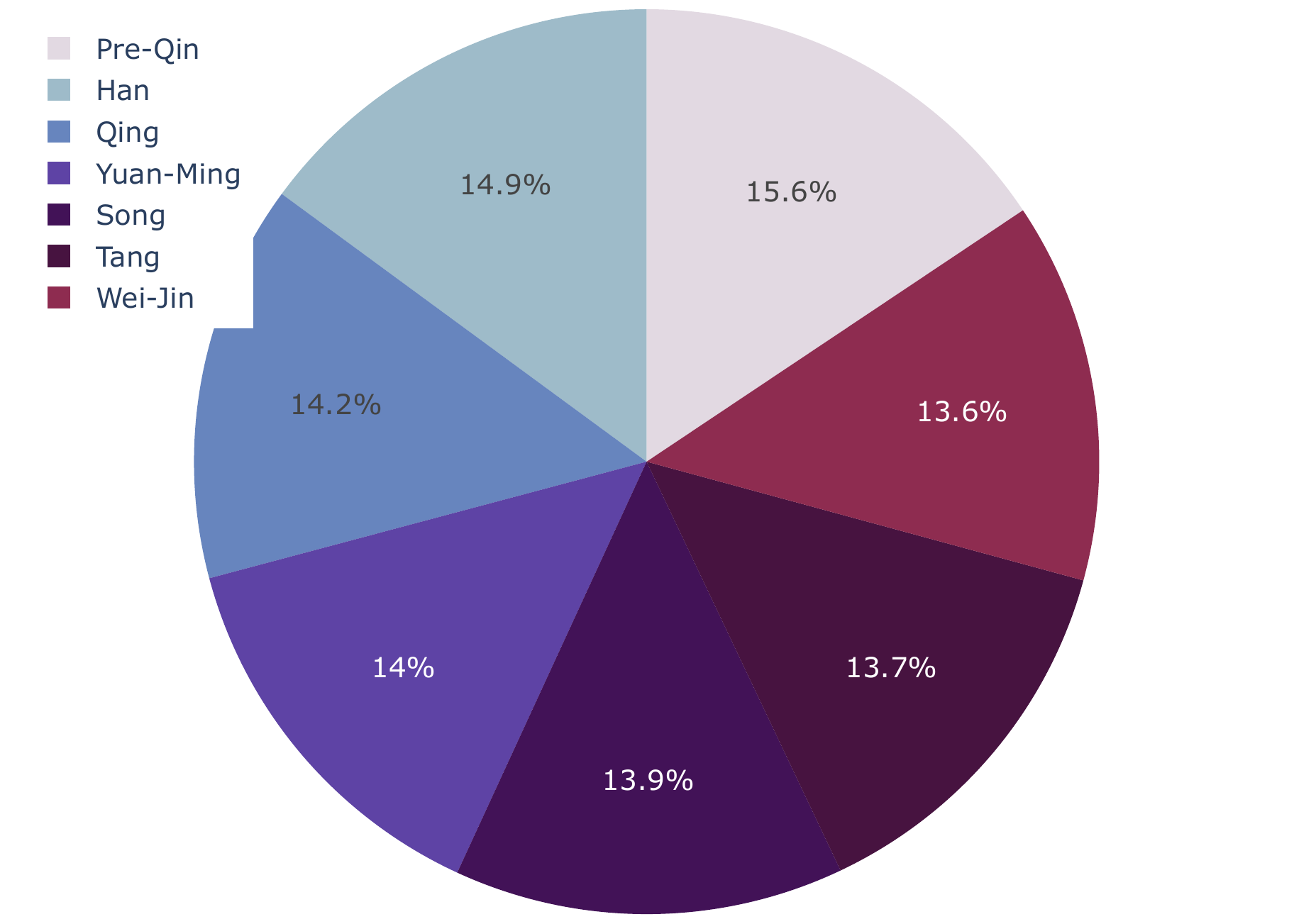}
  \caption{Percentage of Dynasty labels.}
  \label{tlc_label_d:pie}
\end{figure}

\begin{figure}[!t]
	\centering

  \includegraphics[scale=0.4]{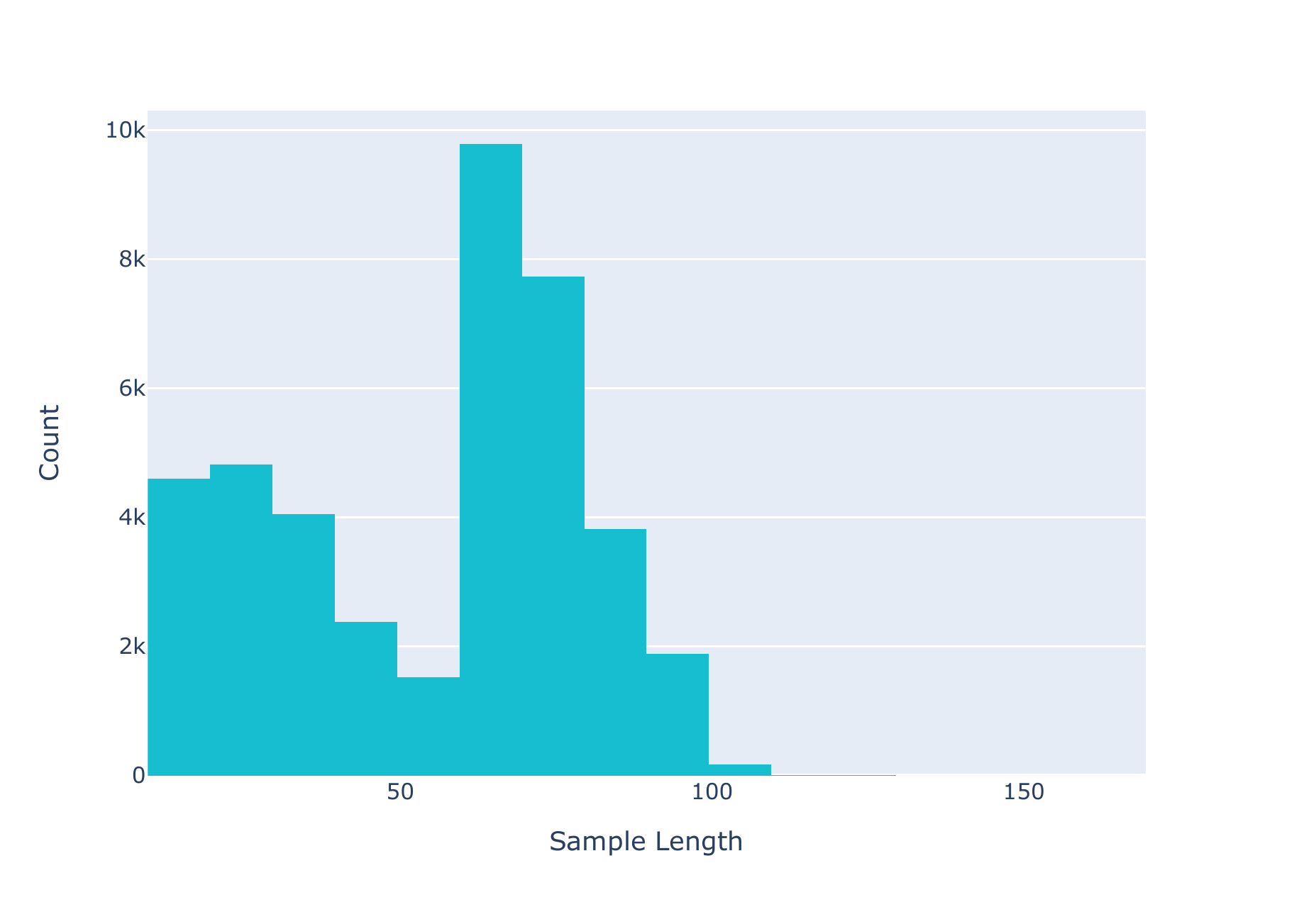}
  \caption{Statistic of Sample Length of TLC dataset.}
  \label{tlc:hist}
\end{figure}

Since ancient books have been handed down over a period of more than 2,000 years, it is a very meaningful and challenging task to identify the writing time of ancient books according to the characteristics of the text. \citet{chang2021time} propose that identifying written time of literature is helpful for understanding works. Being classified according to the period, ancient Chinese is generally divided into ancient (Pre-Qin and Han Dynasty), mid-ancient (Jin Dynasty to Song Dynasty) and late-ancient (Yuan, Ming, Qing Dynasty) \cite{WangLi2004}. Furthermore, in the process of the development of classical Chinese, each dynasty has its own unique characteristics \cite{HYFZS2013}. In such background, we collect about 300 ancient books and famous articles which have exact time of writing, and sample a reasonable number of paragraphs from the texts. This dataset is in TSV format as shown below. The three segments of a sample are Period label, Dynasty label and source text respectively. Statistics are show in Figure~\ref{tlc_label_p:pie}, Figure~\ref{tlc_label_d:pie} Figure~\ref{tlc:hist} and Table~\ref{tlc:Stat}.

\begin{quote}
\textbf{近古} [SEP] \textbf{元明}  [SEP]  主治风痹，筋骨不仁，功与脂同。补虚羸。\\
Indications of wind arthralgia, numbness of the muscles and bones, its power is the same as fat. Make up for weakness.\\
\textbf{中古} [SEP] \textbf{魏晋南北朝} [SEP] 	东观汉记曰：羌什长巩便。然更盖其种也。尚书曰：歼厥渠魁。既已袭汧而馆其县。左氏传曰：凡师轻曰袭。杜预曰：掩其不备。子以眇尔之身，介乎重围之里；率寡弱之众，据十雉之城。\\
Dongguan Han Ji said: Gong Bian, the leader of the Qiang people. But it is another cover. The book of Shang said: Destroy the head of the thief. Has attacked Fianxian and stayed in a hotel. Zuo's biography said: "Any army with light baggage is called 袭." Du Yu said: Attacking who is unprepared. With a small body, you are in the center of the encirclement, leading the weak, and defending the city of ten feet.\\
\textbf{远古} [SEP] \textbf{先秦} [SEP] 齐晏桓子卒，晏婴粗缞斩，苴绖、带、杖，菅屦，食鬻，居倚庐，寝苫、枕草。\\
When father died, Yan Ying wore coarse cloth mourning clothes, made filial piety clothes, belts and walking sticks of coarse linen, wore shoes made of thatch, ate thatch, ate thatch, lived in a leaning hut, and slept on a straw mattress.
\end{quote}

\begin{table}[]
  \centering

  \begin{tabular}{ll}
  \hline
  Total Samples        & 40788 \\ \hline
  Mean Sample Length & 54    \\ \hline
  Min Sample Length  & 11      \\ \hline
  Max Sample Length  & 166    \\ \hline
  \end{tabular}
  \caption{Statistic of Sample Length of TLC dataset.}
  \label{tlc:Stat}

  \end{table}

\subsubsection{Annotation}
\begin{itemize}
    \item Collect books (history books are not included) with a specific written time;
    \item Extract corpora from the books;
    \item Filter out low quality samples, for example, too short or no entities including;
    \item Sample from the result corpora with a specific ratio;
    \item Annotator check;
    \item Final review by experts.
\end{itemize}

\subsection{GJC}
\label{det:gjc}
\subsubsection{Details}
\begin{figure}[!t]
	\centering

  \includegraphics[scale=0.4]{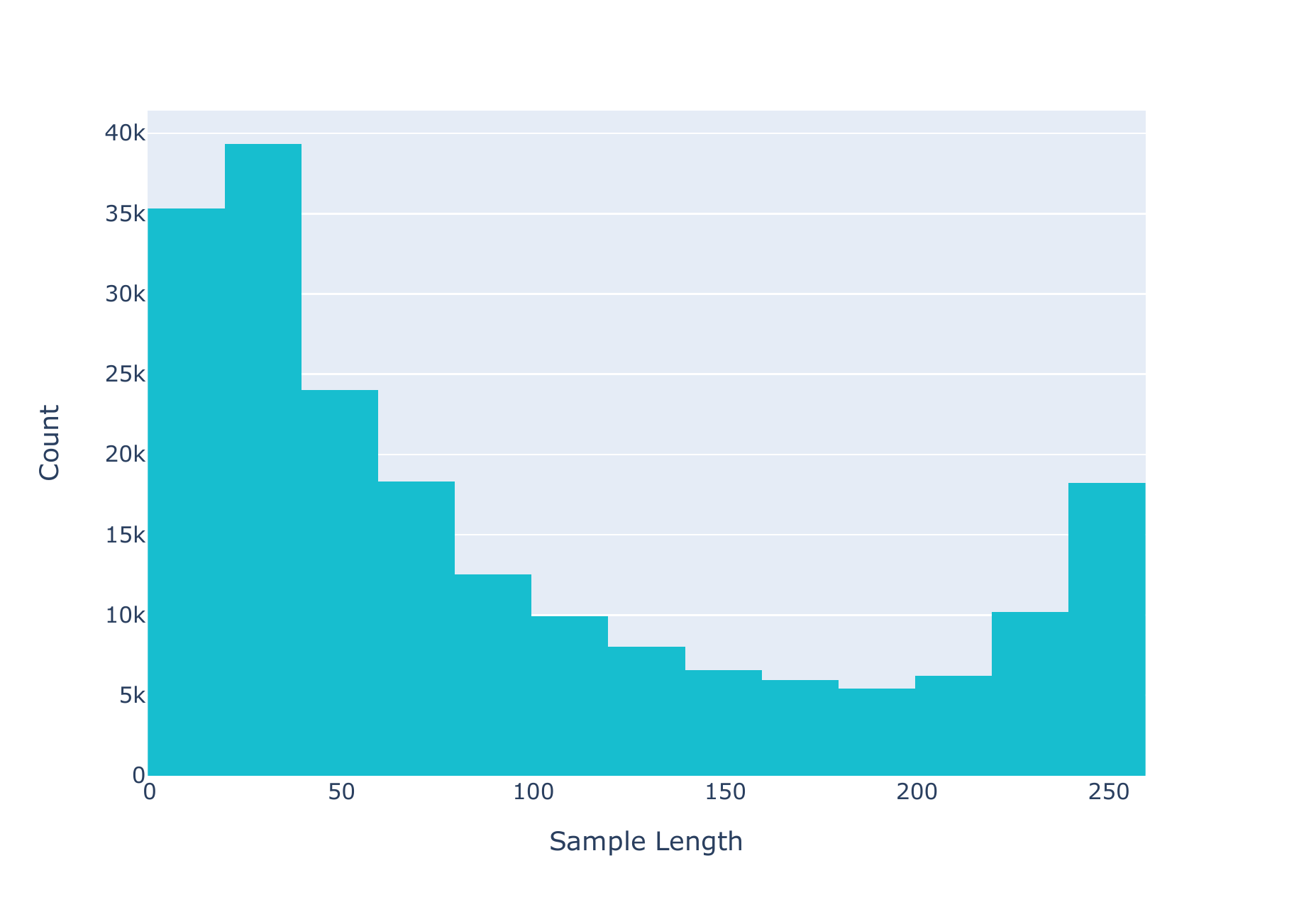}
  \caption{Statistic of Sample Length of GJC dataset.}
  \label{gjc:hist}
\end{figure}

The Si Ku Quan Shu had formed a classification method of four parts of Jing, Shi, Zi, Ji (Confucian classics, historical records, philosophical writings, and miscellaneous works), and 40 categories. This is the authoritative method till now. The largest classical Chinese corpus dataset Daizhige extends the method to 10 collections. Since this corpus is actually the basis of most of classical Chinese NLP research, we apply this method to design our text classification task following CCLUE. This dataset is in text--category format as shown below. Statistics are show in Figure~\ref{gjc:pie} Figure~\ref{gjc:hist} and Table~\ref{gjc:Stat}.

\begin{quote}
  然则世所谓雅乐者，未必如古，而教坊所奏，岂尽为淫声哉？” [SEP] \textbf{艺藏}\\
  However, the elegant music in the society is not necessarily the same as in ancient times, but is the music played by Jiaofang all debauched music? \\
  有丧必求牧师殓，独自入房把门掩。 [SEP] \textbf{子藏}\\
  If there is a funeral, you must find the priest to be buried, entering the room alone and close the door.\\
  “梦幻空花，何劳把捉？得失是非，一时放却。” [SEP] \textbf{佛藏}\\
  "Dreaming of empty flowers, how to take the handle? Do not care about the right and wrong, and put it back at once. ""\\
  羲，乃天皇伏羲氏也。齐驱，即并驾。元始，万有万无之祖号。比肩，并立之义。是足上文比喻也。学者慎毋住相，是即舜何人也，予何人也云尔。 [SEP] \textbf{道藏}\\
  Xi is also called the Emperor Fuxi. “齐驱” means the two marched side by side. “元始”, the ancestor of all things and nothing. "比肩" means to stand side by side. It is enough to describe the above. A scholar must not be too pretentious. This person is what kind of person Shun is, and what kind of person am I.
\end{quote}

\begin{figure}[!t]
	\centering

  \includegraphics[scale=0.4]{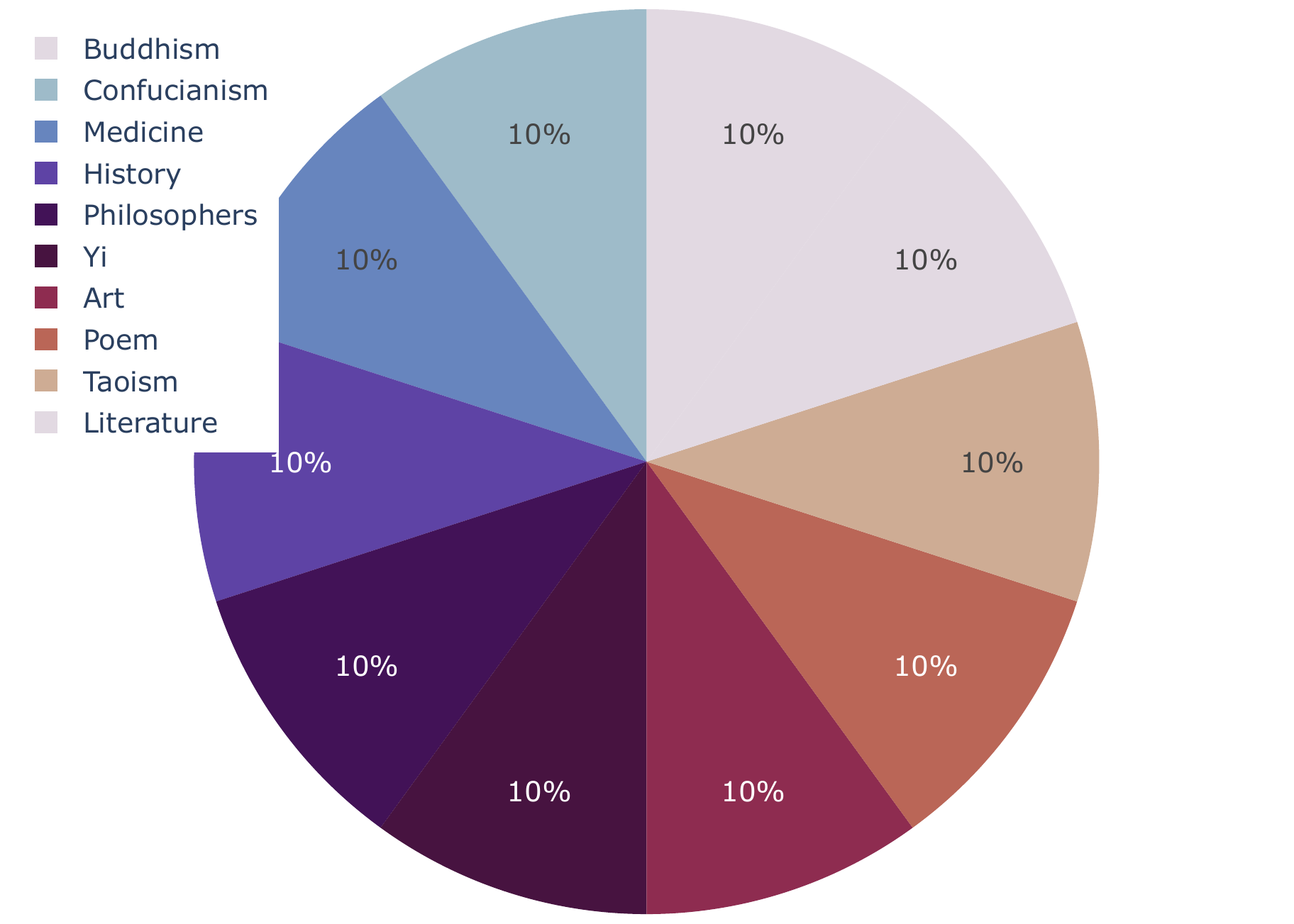}
  \caption{Percentage of labels to be predicted of GJC dataset.}
  \label{gjc:pie}
\end{figure}

\begin{table}[]
  \centering

  \begin{tabular}{ll}
  \hline
  Total Samples        & 200000 \\ \hline
  Mean Sample Length & 92    \\ \hline
  Min Sample Length  & 5      \\ \hline
  Max Sample Length  & 257    \\ \hline
  \end{tabular}
  \caption{Statistic of Sample Length of GJC dataset.}
  \label{gjc:Stat}

  \end{table}

\subsubsection{Annotation}
\begin{itemize}
    \item Collect books from Daizhige corpora;
    \item Extract corpora from the books;
    \item Filter out low quality samples, for example, too short or no entities including;
    \item Sample from the result corpora with a specific ratio;
    \item Annotator check;
    \item Final review by experts.
\end{itemize}

\subsection{FSPC}
\label{det:fspc}

\begin{figure}[!t]
	\centering

  \includegraphics[scale=0.4]{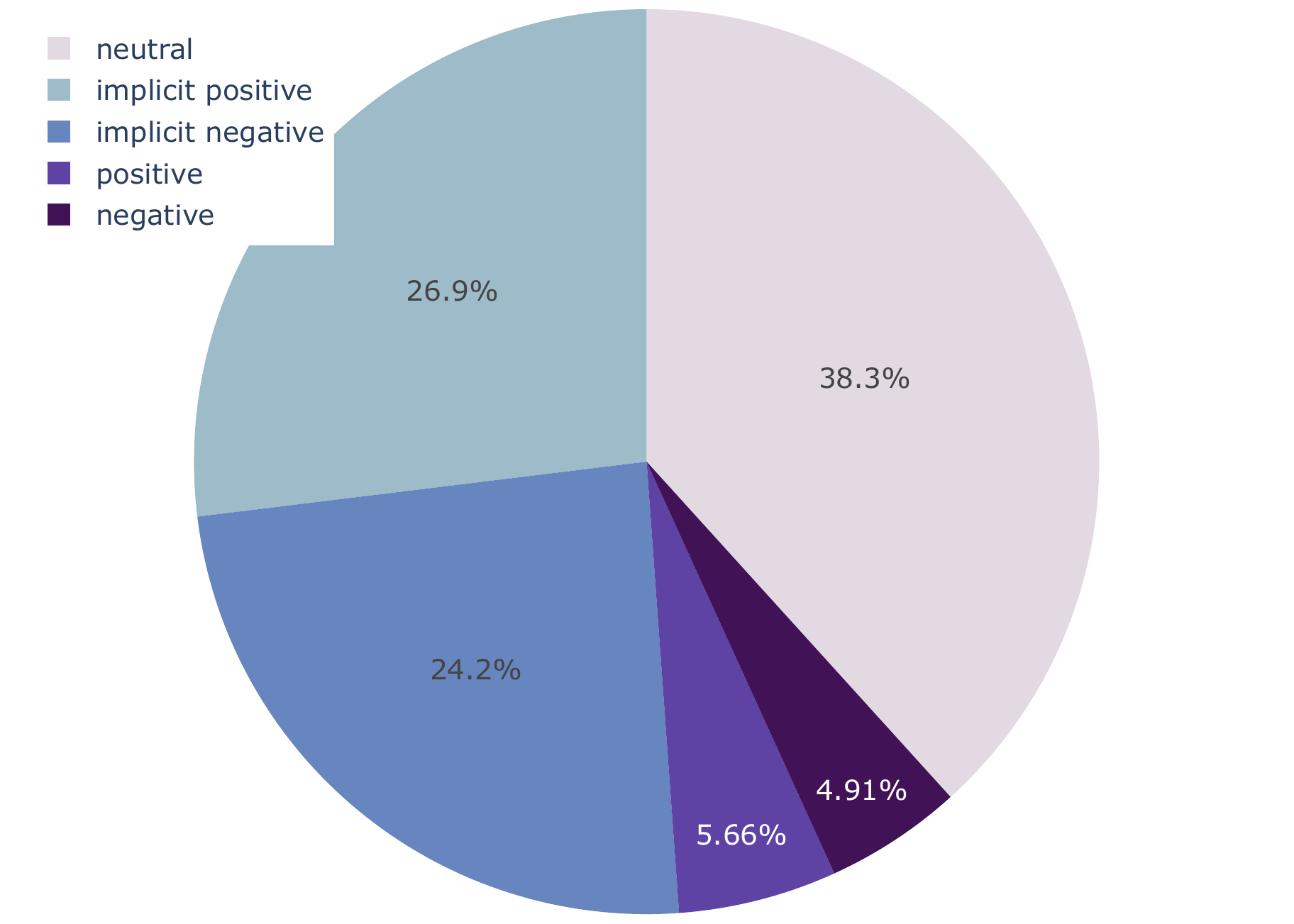}
  \caption{Percentage of labels to be predicted of FSPC dataset.}
  \label{fspc:pie}
\end{figure}

\begin{figure}[!t]
	\centering

  \includegraphics[scale=0.4]{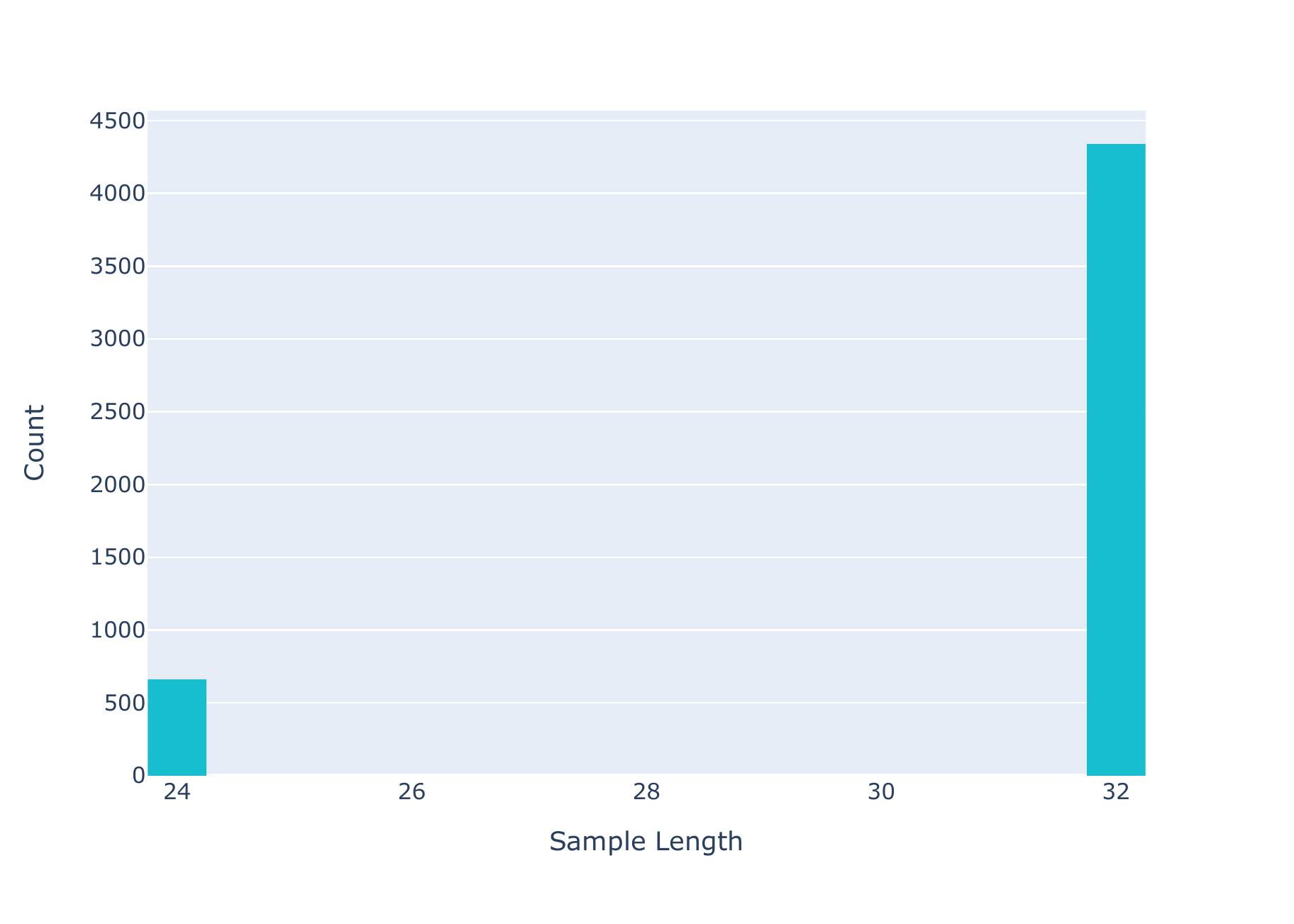}
  \caption{Statistic of Sample Length of FSPC dataset. Note that this dataset only contains five-character quatrains and seven-character quatrains.}
  \label{fspc:hist}
\end{figure}

This dataset is in JSON format as shown below. The sentiment labels are of five specifications which shift from negative to positive. Statistics are show in Figure~\ref{fspc:pie} Figure~\ref{fspc:hist}.

\begin{quote}
  \{\\
    "poet": "范仲淹",\\
    "poem": "静映寒林晚未芳|人人欲看寿阳妆|玉颜须傍韶春笑|莫斗严风与恶霜",\\
    Quietly reflecting the cold, the forest is late and not fragrant | Everyone wants to see Shouyang makeup | Jade face must be close to spring smile | Do not fight with strong wind and cold frost.\\
    "dynasty": "宋",\\
    "sentiments": \{\\
     "holistic": "implicit positive",\\
     "line1": "implicit positive",\\
     "line2": "neutral",\\
     "line3": "implicit positive",\\
     "line4": "neutral"\\
    \},\\
    "title": "和提刑赵学士探梅三绝"\\
   \},\\
   \{\\
    "poet": "王维",\\
    "poem": "独在异乡为异客|每逢佳节倍思亲|遥知兄弟登高处|遍插茱萸少一人",\\
    Being alone and a stranger in a foreign land | I miss my relatives every time during the festival | I know my brothers climb a high place from afar | Everyone plant cornel all over the place but me\\
    "dynasty": "唐",\\
    "sentiments": \{\\
     "holistic": "implicit negative",\\
     "line1": "implicit negative",\\
     "line2": "implicit negative",\\
     "line3": "neutral",\\
     "line4": "implicit negative"\\
    \},	
\end{quote}

\subsection{Xuci}
\label{det:xuci}
\subsubsection{Details}
Function words (Xu ci in Chinese) have no real meaning and generally cannot be used as a single sentence element \cite{Liu1995XUCI}. They are very important in classical Chinese but easily confused. Relevant topics are part of the basic knowledge for Chinese students which appears in the college entrance exam every year. We collect sentence pairs with function words from examination papers with help of middle school teachers to construct this dataset.

\begin{figure}[!t]
	\centering

  \includegraphics[scale=0.4]{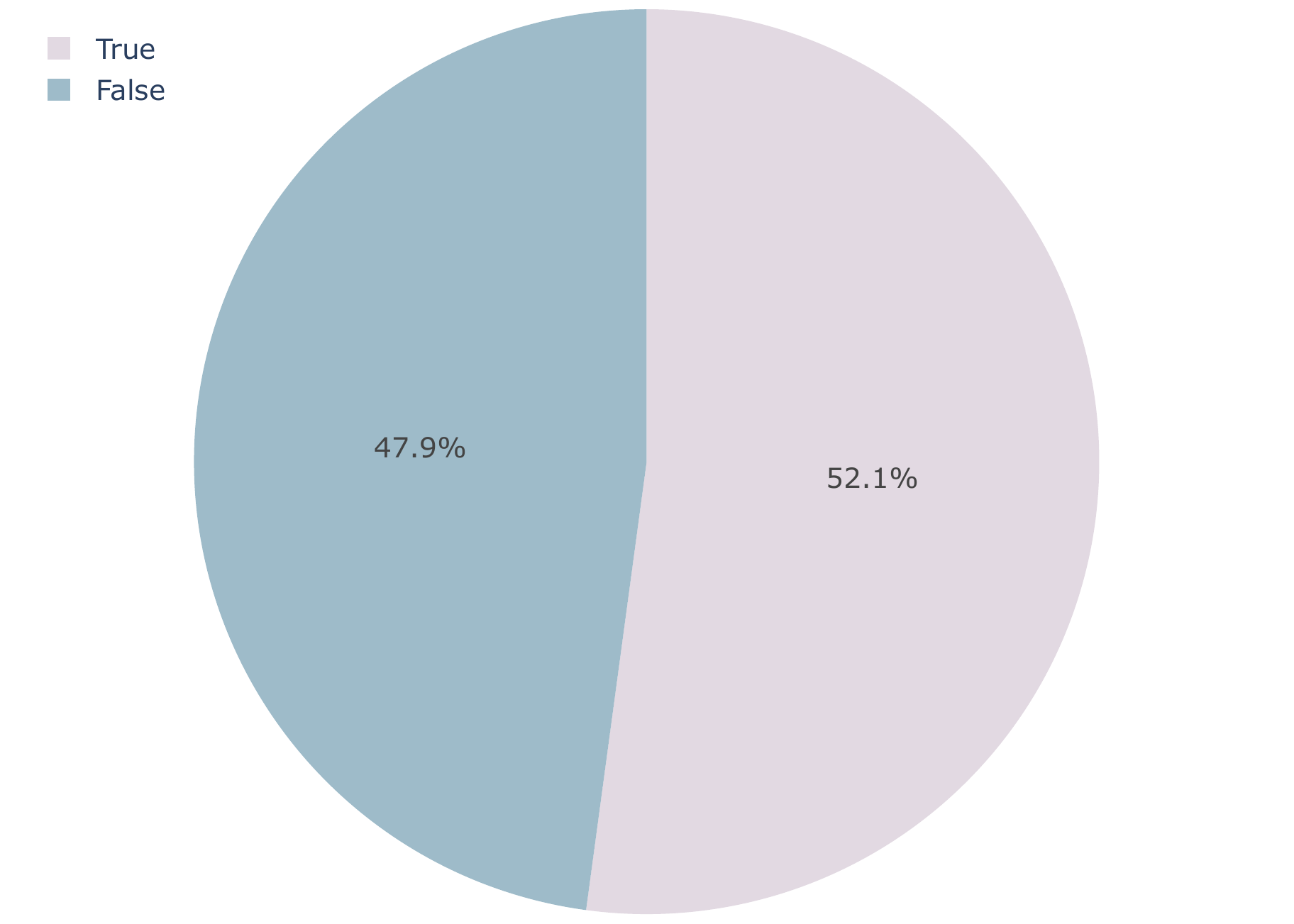}
  \caption{Percentage of labels to be predicted of Xuci dataset.}
  \label{xuci:pie}
\end{figure}

\begin{figure}[!t]
	\centering

  \includegraphics[scale=0.4]{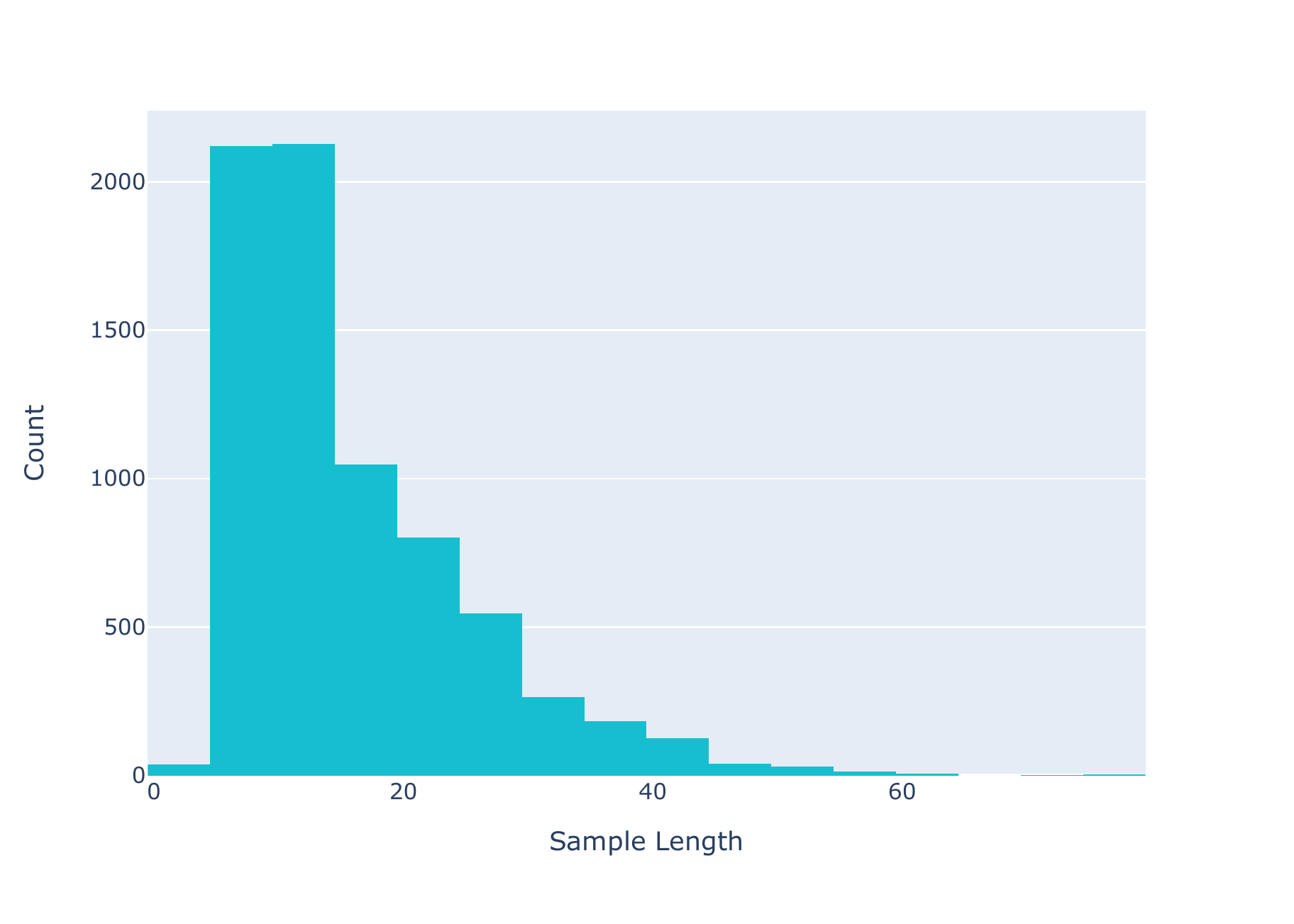}
  \caption{Statistic of Sample Length of Xuci dataset.}
  \label{xuci:hist}
\end{figure}

This dataset is in TSV format. Statistics are show in Figure~\ref{xuci:pie} Figure~\ref{xuci:hist} and Table~\ref{xuci:Stat}.

\begin{quote}
  使夫邪污之气无由得接焉。[SEP] 复驾言兮焉求。[SEP] 10, 10 [SEP] 4, 4 [SEP] f\\
  so that there is no way for those evil and filthy atmospheres to reach them. [SEP]What am I driving for?\\
  上官令民送牛羊之陕西。[SEP] 久之，举于朝。[SEP] 7, 7 [SEP] 1, 1 [SEP] f\\
  The superior commander sent cattle and sheep to Shaanxi. [SEP] After a long time, he was recommended to the court.\\
  容与乎阳林，流眄乎洛川。[SEP] 她也曾近乎撒娇似地问过他。[SEP] 2, 2 [SEP] 4, 4 [SEP] t\\
However, he calmly left the sun and looked at Luochuan with vast water. [SEP] She had also asked him almost coquettishly.
\end{quote}

\begin{table}[]
  \centering

  \begin{tabular}{ll}
  \hline
  Total Samples        & 7350 \\ \hline
  Mean Sample Length & 16    \\ \hline
  Min Sample Length  & 3      \\ \hline
  Max Sample Length  & 79    \\ \hline
  \end{tabular}
  \caption{Statistic of Sample Length of Xuci dataset.}
  \label{xuci:Stat}

  \end{table}

\subsubsection{Annotation}
\begin{itemize}
\item Prepossess test papers, including OCR, layout parser and then copy all word comparison problems;
\item Annotate, including filter the sub-problems for which suitable for machine learning and correct misspelling, etc.;
\item Cross double-check between annotators;
\item Final review by experts.
\end{itemize}


\subsection{IRC}
\label{det:irc}
\subsubsection{Details}
\begin{figure}[!t]
	\centering
  \includegraphics[width=1\linewidth]{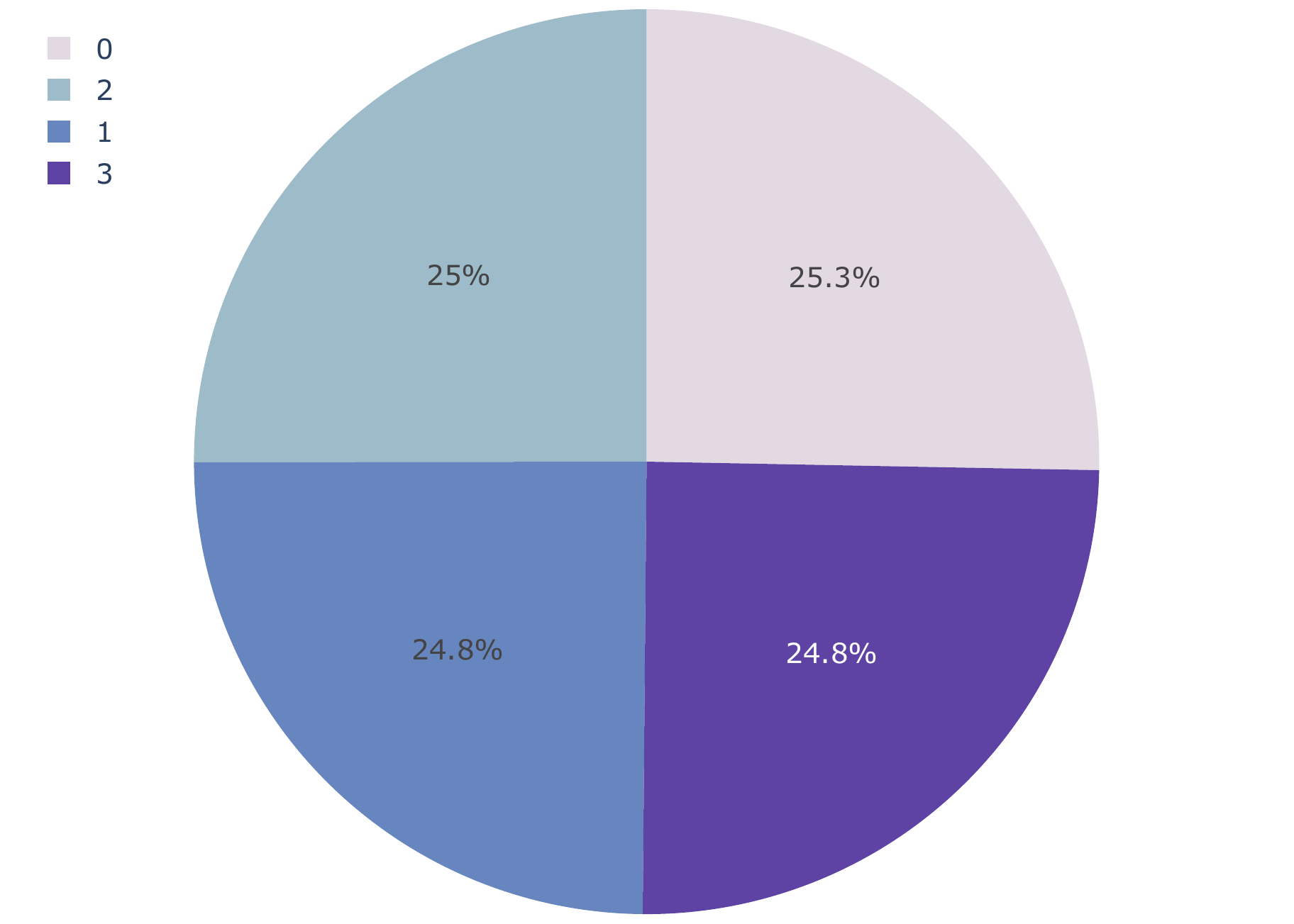}
  \caption{Percentage of labels to be predicted of IRC dataset.}
  \label{irc:pie}
\end{figure}

\begin{figure}[!t]
	\centering
  \includegraphics[width=1\linewidth]{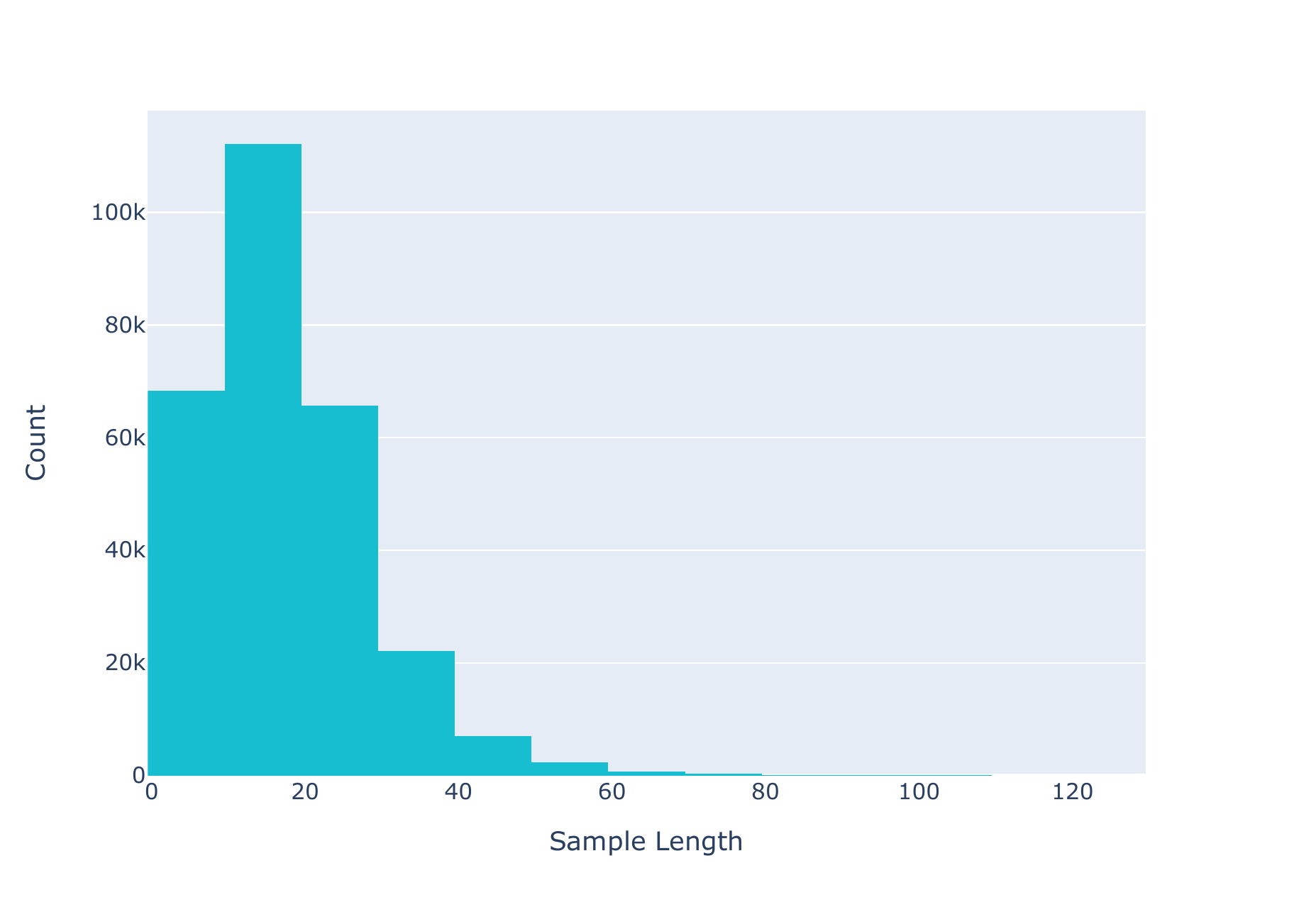}
  \caption{Statistic of Sample Length of IRC dataset.}
  \label{irc:hist}
\end{figure}

Idiom is one of the major features of Chinese culture. Most of the idioms are long-standing fixed phrases, derived from ancient classics or writings, historical stories, and oral stories. For idiom comprehension, there are other tasks 
 \cite{zheng2019chid} ready. However, they are mainly aiming to test modern Chinese texts with idioms. To focus on classical Chinese, we implement this task。
This dataset is in JSON format. Every sample consist of four fields which are "idiom", "options", "label" and "origin". The ground truth "label" is best fit of the four options. Statistics are show in Figure~\ref{irc:pie} Figure~\ref{irc:hist} and Table~\ref{irc:Stat}.
\begin{quote}
  \{
    "idiom": "眼去眉来",\\
    eye to eyebrow\\
    "options": {[}\\
     "火烧到眉毛。比喻事到眼前，非常急迫。",\\
     The fire burned to the eyebrows. The metaphor is very urgent.\\
     "形容事情已到眼前，情势十分紧迫。",\\
     Describe the matter has come to the front, the situation is very urgent.\\
     "原指眼前见到的。后形容用眉眼传情。",\\
     It meant what was seen. After describing the use of eyebrows teasing.\\
     "形容眉眼含情示意的神态。"\\
     Describe the expression of the eyebrows showing affection.\\
    {]},
    "label": 2,
    "origin": "落日苍茫，风才定，片帆无力。还记得眉来眼去，水光山色。"\\
    The setting sun is vast, the wind is fixed, and the sails are weak. I still remember the frowning, the water and the mountains.
   \},\\
\end{quote}

\begin{table}[]
  \centering

  \begin{tabular}{l|l|l|l|l|l|l}
    \hline
                         & Idiom & Origin & 1 & 2 & 3 & 4 \\ \hline
    Total        & \multicolumn{6}{c}{46471}                              \\ \hline
    Mean & -     & 19     & 18      & 18      & 16      & 16      \\ \hline
    Min  & 4     & 5      & 3       & 3       & 4       & 4       \\ \hline
    Max  & 16    & 121    & 76      & 80      & 75      & 76      \\ \hline
    \end{tabular}
  \caption{Statistic of Sample Length of IRC dataset.}
  \label{irc:Stat}

  \end{table}
\subsubsection{Annotation}
\begin{itemize}
\item Prepossess test papers, including OCR, layout parser and then copy all idiom comprehension problems;
\item Study the idiom dictionary to extract idiom comprehension problems;
\item Annotate, including filter the sub-problems for which suitable for machine learning and correct misspelling, etc.;
\item Cross double-check between annotators;
\item Final review by experts.
\end{itemize}

\subsection{WYWMT}
\label{det:wywmt}
\subsubsection{Details}
Classical Chinese is a very concise written language, so it’s not easy for everyone to understand. Scholars often translate classical Chinese into modern Chinese with notations to make it easier for people to read. We consider it as an in-language translation or rewriting task because the source and target could share the same vocabulary and some semantic features. 

This dataset is filtered and calibrated from hundreds of translated classical Chinese books collected from multiple channels. Since allusions and quotations appear frequently in classical Chinese, and these references may have a time span of thousands of years, it’s not easy to construct a very well-established dataset by ourselves.

\begin{figure}[H]
  \centering
  \includegraphics[width=1\linewidth]{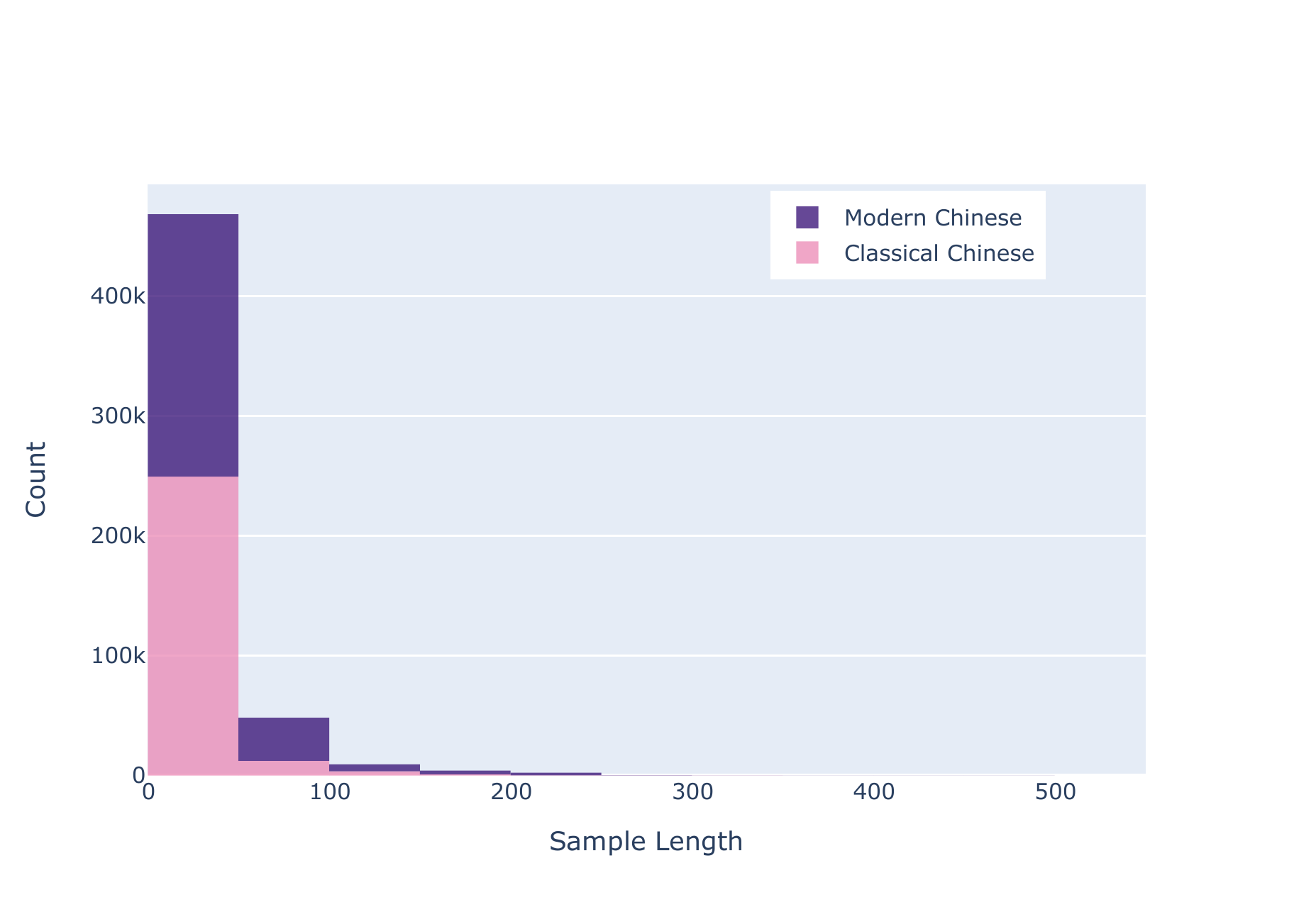}
  \caption{Statistic of Classical Sample Length of WYWMT dataset.}
  \label{mt:histc}
\end{figure}

This dataset is in sentence pair TSV format. Samples are represented as "source" and "reference" segment which are separated by "tab". Statistics are show in Figure~\ref{mt:histc} and Table~\ref{mt:Stat}.

\begin{quote}
  共四里，又越一冈脊而下，其脊高不及高井之半，而实为西北来过脊以趋清秀者也。[SEP] 共四里，又越一道冈脊后下走，这个冈脊高处不到高井的一半，但实际上是从西北前来趋向清秀山的延伸而过的山脊。\\
  After a total of four miles, we went down after another ridge. This ridge was less than half of the height of Gaojing, but it was actually a ridge extending from the northwest towards Qingxiu Mountain.\\
  读性理书时，则杂以诗文各集，以歧其趋。[SEP] 在读性理书的时候，又掺杂写诗文，走了岔路。\\
  When I read books about ethics, I mixed it with writing poetry, so I went to a wrong road.
\end{quote}


\begin{table}[H]
  \centering

  \begin{adjustbox}{width=0.48\textwidth}
  \begin{tabular}{l|c|c}
    \hline
                         & \textbf{Classical}     & \textbf{Modern}    \\ \hline
    Total Samples        & \multicolumn{2}{c}{46471} \\ \hline
    Mean Sample Length & 23            & 37        \\ \hline
    Min Sample Length  & 5             & 5         \\ \hline
    Max Sample Length  & 381           & 508       \\ \hline
    \end{tabular}
    \end{adjustbox}
  \caption{Statistic of Sample Length of WYWMT dataset.}
  \label{mt:Stat}

  \end{table}

\subsubsection{Annotation}
\begin{itemize}
\item Crawl classical Chinese articles with modern translation from all channels, note that these articles are open and free for everyone;
\item Text align;
\item Split long paragraphs into shorter sentences;
\item Filter low quality examples;
\item Final review by experts.
\end{itemize}

\section{Details of human evaluation}
\label{app:human}
To obtain a more reliable evaluation of model performance, we chose students majoring in Classical Chinese to provide an upper bound score. This sets a higher value for the benchmark and helps researchers improve their models. The students were given access only to the train-set during the test, without any additional tools. For different tasks, set of the test examples is different. See Table~\ref{tab:human_details}.
\begin{table}[]
\begin{adjustbox}{width=0.48\textwidth}
\begin{tabular}{ccccccc}
Task  & Testers & Examples & Tester 1 & Tester 2 & Tester 3 & Average \\ \hline
PUNC  & 3       & 100      & 90.2     & 93.2     & 93.8     & 92.4    \\
GLNER & 3       & 100      & 94.4     & 93.1     & 95.4     & 94.3    \\
GJC   & 3       & 100      & 89.0     & 88.0     & 94.0       & 90.3    \\
FSPC  & 3       & 20       & 83.0     & 77.0     & 80.0     & 80.0    \\
TLC   & 3       & 100      & 87.0     & 86.0     & 94.0     & 89.0    \\
Xuci  & 3       & 100      & 86.0     & 83.0     & 87.0     & 85.3    \\
WYWRC & 3       & 20       & 75.0     & 80.0     & 85.0     & 80.0    \\
IRC   & 3       & 100      & 90.0     & 93.0     & 94.0       & 92.3    \\
WYWMT & 3       & 20       & 43.8     & 45.2     & 47.8     & 45.6   
\end{tabular}
\end{adjustbox}
\caption{Details of human evaluation. Tester 1 is a college freshman students. Tester 2 is a third-year university student. Tester 3 is a graduate students.}
\label{tab:human_details}
\end{table}

\section{Details of Models Evaluated}
\label{app:det_models}
In this section, we present the details of pre-trained language models we used, including guwenbert-base, guwenbert-large, guwenbert-base-fs, roberta-classical-chinese-base-char, roberta-classical-chinese-large-char, SikuBERT, SikuRoBERTa, DeBERTa-base and RoBERTa-wwm-ext. As shown in~\ref{tab:model_details}, the masking, scale, corpus, vocabulary and parameter initialization are different in each pre-trained language model.

\begin{table*}[t]
\centering

\begin{adjustbox}{width=1\textwidth}
\begin{tabular}{lcccccc}
\hline
\textbf{Model} &
  \textbf{Masking} &
  \textbf{Scale} &
  \textbf{Corpus} &
  \textbf{Optimizer} &
  \textbf{Vocabulary} &
  \textbf{Init.} \\ \hline
guwenbert-base    & WWM    & base  & DaiZhiGe       & AdamW & 23292 & RoBERTa-wwm-ext  \\
guwenbert-large   & WWM    & large & DaiZhiGe       & AdamW & 23292 & RoBERTa-wwm-ext  \\
guwenbert-base-fs & WWM    & base  & DaiZhiGe       & AdamW & 23292 & Scrach Classical \\
\begin{tabular}[c]{@{}c@{}}roberta-classical-\\  chinese-base-char\end{tabular} &
  Mask &
  base &
  DaiZhiGe &
  AdamW &
  26318 &
  guwenbert-base \\
\begin{tabular}[c]{@{}c@{}}roberta-calssical-\\  chinese-large-char\end{tabular} &
  Mask &
  large &
  DaiZhiGe &
  AdamW &
  26318 &
  guwenbert-large \\
SikuBERT          & Mask   & base  & Sikuquanshu    & AdamW & 29791 & Scrach Classical \\
SikuRoBERTa       & Mask   & base  & Sikuquanshu    & AdamW & 29791 & Scrach Classical \\
DeBERTa-base      & n-gram & base  & DaiZhiGe       & AdamW & 22669 & Scrach Classical \\
RoBERTa-wwm-ext   & WWM    & base  & Chinese Corpus & AdamW & 21128 & Scrach Modern    \\ \hline
\end{tabular}
\end{adjustbox}

\caption{Parameters for pretraining of collected models.}
\label{tab:model_details}
\end{table*}

\section{Hyper-parameters for fine-tuning}
\label{app:hp}
As shown in Table~\ref{tab:hp}, we present the hyper-parameters applied in fine-tuning. For different scale of pre-trained language model, we set different learning rates. In large scale, we set learning rates with 5e-6, 8e-6, 9e-16 and 1e-5. In base scale, we set learning rates from 1e-5 to 5e-5. We set warmup to 0.1, maximum epochs to 10. For Adam, we set $\epsilon$ to 1e-6, $\beta_1$ and $\beta_2$ to 0.9 and 0.999 respectively. Meanwhile, we use linear for LR decay and set weight decay to 0.01.

\begin{table}[H]

\begin{adjustbox}{width=0.48\textwidth}
\begin{tabular}[b]{lll}
\hline
\textbf{Hyper-parameter} & \textbf{Large scale}       & \textbf{Base scale} \\
\hline
Dropout    & \{0,0.1,0.15\}             & \{0,0.1,0.15\}      \\
Warmup        & 0.1                        & 0.1                 \\
Learning Rates           & \{5e-6, 8e-6, 9e-6, 1e-5\} & \{1e-5 to 5e-5\}    \\
Batch Size               & \{16,32,48,64\}            & \{16,32,48,64\}     \\
Weight Decay             & 0.01                       & 0.01                \\
Maximum Epochs  & 10                         & 10                  \\
LR Decay      & Linear                     & Linear              \\
Adam $\epsilon$                   & 1e-6                       & 1e-6                \\
Adam $\beta_1$                & 0.9                        & 0.9                 \\
Adam $\beta_2$                & 0.999                      & 0.999               \\
Gradient Clipping        & 1.0                        & 1.0                \\
\hline
\end{tabular}
\end{adjustbox}

\caption{Hyper-parameters for fine-tuning.}
\label{tab:hp}
\end{table}

\section{Leader-board}
\label{leaderboard}
Following other benchmark leaderboard, this leaderboard is designed containing an overall list and several sub-lists. Metrics of the tasks are listed in Table~\ref{tab:overview}. Since most of the tasks are common tasks, we did not design more metrics for them. For the sequence-to-sequence task of WYWMT, due to the shortcomings of various indicators, we calculate several metrics for the prediction results to make that clearer (\ref{tab:resultsMT}).
\begin{figure}[!t]
\centering
  \includegraphics[width=1\linewidth]{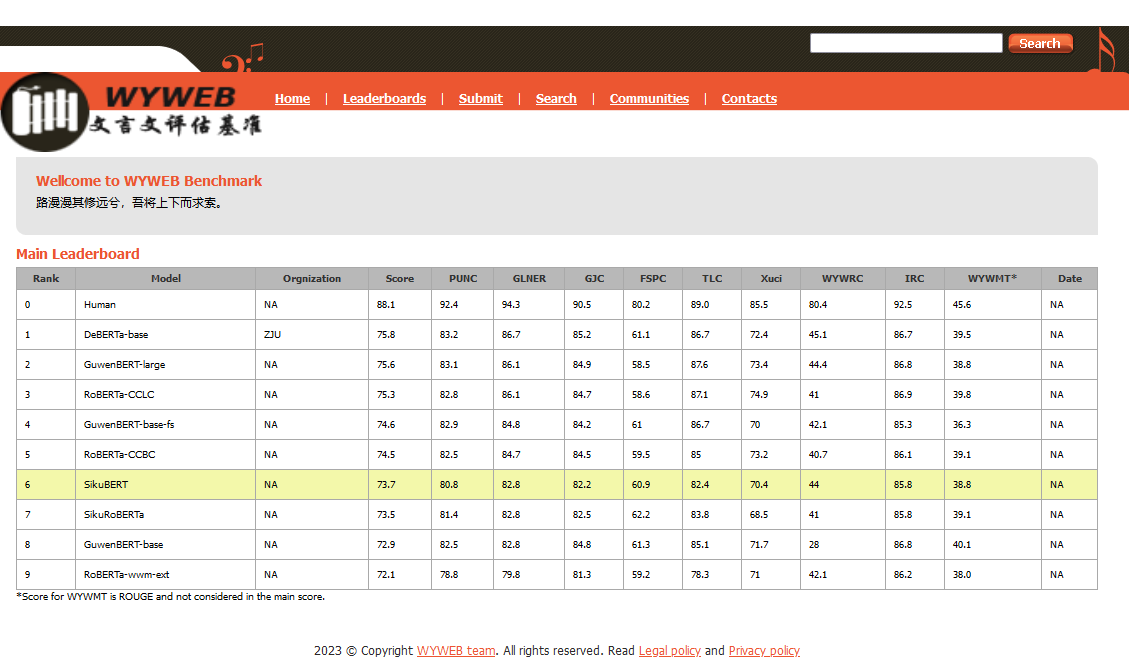}
  \caption{Screen-shot of home page of the leader-board.}
  \label{leaderboard:main}
\end{figure}
\begin{figure}[!t]
\centering
  \includegraphics[width=1\linewidth]{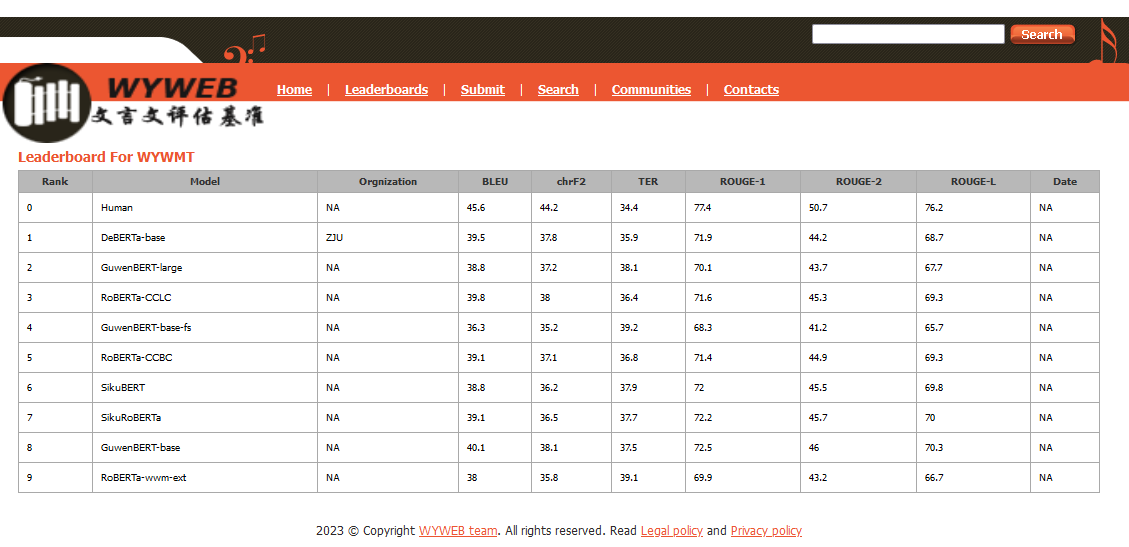}
  \caption{Screen-shot of WYWMT rank page of the leader-board.}
  \label{leaderboard:WYWMT}
\end{figure}
\end{CJK*}
\end{document}